\pdfoutput=1

\documentclass[11pt]{article}

\usepackage[]{acl}

\usepackage{times}
\usepackage{latexsym}

\usepackage[T1]{fontenc}

\usepackage[utf8]{inputenc}

\usepackage{microtype}
\usepackage{graphicx}
\usepackage{todonotes}

\usepackage{booktabs}
\usepackage{comment}
\usepackage{xcolor}         
\usepackage{colortbl}         
\usepackage{amssymb}
\usepackage{hyperref}      
\usepackage{colortbl}  
\usepackage{url}
\usepackage{gb4e}
\noautomath

\newcommand{\word}[1] {`\textit{#1}'}
\newcommand{\sense}[1] {`\textsc{{#1}}'}

\newcommand{\mail}[1] {\url{\texttt{#1}}}

\title{Interpretable Word Sense Representations via Definition Generation:\\The Case of Semantic Change Analysis}

\newcommand{\amsterdam}[0]{$^{\triangleleft}$}
\newcommand{\oslo}[0]{$^{\diamond}$}

\author{Mario Giulianelli\amsterdam,\ \ Iris Luden\amsterdam,\ \ Raquel Fern\'andez\amsterdam,\ \ Andrey Kutuzov\oslo \\
\amsterdam University of Amsterdam\ \ \oslo University of Oslo\\
\url{m.giulianelli@uva.nl},\ \ \url{irisluden@gmail.com},\\ \url{raquel.fernandez@uva.nl},\ \ \url{andreku@ifi.uio.no}
}
\begin{document}
\maketitle

\begin{abstract}

We propose using automatically generated natural language definitions of contextualised word usages as interpretable word and word sense representations.
Given a collection of usage examples for a target word, and the corresponding data-driven usage clusters (i.e., word senses), a definition is generated for each usage with a specialised Flan-T5 language model, and the most prototypical definition in a usage cluster is chosen as the sense label. 
We demonstrate how the resulting sense labels can make existing approaches to semantic change analysis more interpretable, and how they can allow users---historical linguists, lexicographers, or social scientists---to explore and intuitively explain diachronic trajectories of word meaning. Semantic change analysis is only one of many possible applications of the `definitions as representations' paradigm. Beyond being human-readable, contextualised definitions also outperform token or usage sentence embeddings in word-in-context semantic similarity judgements, making them a new promising type of lexical representation for NLP.\looseness-1
\end{abstract}


\section{Introduction} 
\label{sec:intro}
\begin{table*}[ht]
    \centering
    \resizebox{\textwidth}{!}{%
    \begin{tabular}{p{220px}|c|p{180px}}
    \toprule
    \textbf{Usage example} & \textbf{Target word} & \textbf{Generated definition} \\
    \midrule
    `about half of the soldiers in our rifle platoons were \textbf{draftees} whom we had trained for about six weeks' & \textbf{draftee} & \sense{a person who is being enlisted in the armed forces} \\
    \bottomrule
    \end{tabular}
    }
    \caption{An example of a definition generated by our fine-tuned Flan-T5 XL. The model is prompted with the usage example, post-fixed with the phrase \textit{`What is the definition of draftee?'}}
    \label{tab:draftee}
\end{table*}

Accurate semantic understanding in language technologies is
typically powered by distributional word representations and pre-trained language models (LMs). Due to their subsymbolic nature, however, such methods lack in explainability and interpretability, leading to insufficient trust in end users. 
%
An example application which requires capturing word meaning with its nuanced context-determined modulations is \textit{lexical semantic change} analysis, a task which consists in detecting whether a word's meaning has changed over time, for example by acquiring or losing a sense.
Modern semantic change detection systems rely on static and contextualised word representations, LM-based lexical replacement, grammatical profiles, supervised word sense and word-in-context disambiguation \cite{kutuzov-etal-2018-diachronic,tahmasebia2021survey}.
But the main potential end users of these technologies---historical linguists, lexicographers, and social scientists---are still somewhat reluctant to adopt them precisely because of their lack of explanatory power. Lexicographers, for instance, are not satisfied with detecting that a word has or hasn't changed its meaning over the last ten years; they want descriptions of old and new senses in human-readable form, possibly accompanied by additional layers of explanation, e.g., specifying the type of semantic change (such as broadening, narrowing, and metaphorisation) the word has undergone.

Our work is an attempt to bridge the gap between computational tools for semantic understanding and their users. 
We propose to replace black-box contextualised token embeddings produced by large LMs with a new type of interpretable lexical semantic representation: automatically generated \textit{contextualised word definitions} \cite{gardner2022definition}. 
In this paradigm, the usage of the word \word{apple} in the sentence \word{She tasted a fresh green apple} is represented not with a dense high-dimensional vector but with the context-dependent natural language definition \sense{edible fruit}. 
With an extended case study on lexical semantic change analysis, we show that moving to the more abstract meaning space of definitions allows practitioners to obtain explainable predictions from computational systems, while leading to superior performance on semantic change benchmarks compared to state-of-the-art token-based approaches.

This paper makes the following contributions.\footnote{All the code we used can be found at \url{https://github.com/ltgoslo/definition_modeling}.}
\vspace{-0.4em}
\begin{enumerate}
\itemsep=0em

    \item We show that word definitions automatically generated with a specialised language model, fine-tuned for this purpose, can serve as interpretable representations for polysemous words (§\ref{sec:pairwise}). Pairwise usage similarities between contextualised definitions approximate human semantic similarity judgements better than similarities between usage-based word and sentence embeddings. 
    \item We present a method to obtain \textit{word sense representations} by labelling data-driven clusters of word usages with sense definitions, and collect human judgements of definition quality to evaluate these representations~(§\ref{sec:labels}). We find that sense labels produced by retrieving the most prototypical contextualised word definition within a group of usages consistently outperform labels produced by selecting the most prototypical token embedding.
    \item Using sense labels obtained via definition generation, we create maps that describe diachronic relations between the senses of a target word. We then demonstrate how these \textit{diachronic maps} can be used to explain meaning changes observed in text corpora and to find inconsistencies in data-driven groupings of word usages within existing lexical semantic resources (§\ref{sec:explain}).
\end{enumerate}

\section{Related Work}
\label{sec:background}

\subsection{Definition Modelling}
\label{subsec:background-def-generation}

The task of generating human-readable word definitions, as found in dictionaries, is commonly referred to as \textit{definition modelling} or \textit{definition generation} \citep[for a review, see][]{gardner2022definition}. 
The original motivation for this task has been the interpretation, analysis, and evaluation of word embedding spaces. Definition generation systems, however, also have practical applications in lexicography, language acquisition, sociolinguistics, and within NLP~\cite{bevilacqua-etal-2020-generationary}.
%
The task was initially formulated as the generation of a natural language definition given an embedding---a single distributional representation---of the target word, or \textit{definiendum}~\citep{noraset2017definition}.
Word meaning, however, varies according to the context in which a word is used. This is particularly true for polysemous words, which can be defined in multiple, potentially very different ways depending on their context. The first formulation of definition modelling was therefore soon replaced by the task of generating a contextually appropriate word definition given a target word embedding and an example usage~\cite{gadetsky-etal-2018-conditional,mickus-etal-2022-semeval}.
When the end goal is not the evaluation of embedding spaces, generating definitions from vector representations is still not the most natural formulation of definition modelling. \citet{ni-wang-2017-learning} and \citet{mickus-etal-2019-mark} treat the task as a sequence-to-sequence problem: given an input sequence with a highlighted word, generate a contextually appropriate definition. In the current work, we follow this approach. Table~\ref{tab:draftee} shows an example of a contextualised definition generated by our model (see §\ref{sec:method}) for the English word \word{draftee}.\looseness-1

\paragraph{Methods}
Approaches that address this last formulation of the task are typically based on a pre-trained language model deployed on the definienda of interest in a natural language generation (NLG) setup~\cite{bevilacqua-etal-2020-generationary}. 
Generated definitions can be further improved by regulating their degree of specificity via specialised LM modules~\citep{huang-etal-2021-definition}, by adjusting their level of complexity using contrastive learning training objectives~\citep{august-etal-2022-generating}, or by supplementing them with definitional sentences extracted directly from a domain-specific corpus~\citep{huang-etal-2022-jargon}. We will compare our results to the specificity-tuned T5-based text generator proposed by \citet{huang-etal-2021-definition}.\looseness-1

\paragraph{Evaluation} Generated definitions are typically evaluated with standard NLG metrics such as BLEU, NIST, ROUGE-L, METEOR or MoverScore \cite[e.g.,][]{huang-etal-2021-definition,mickus-etal-2022-semeval}, using precision@k on a definition retrieval task~\citep{bevilacqua-etal-2020-generationary}, or measuring semantic similarity between sentence embeddings obtained for the reference and the generated definition~\citep{kong-etal-2022-multitasking}. Because reference-based methods are inherently flawed \cite[for a discussion, see][]{mickus-etal-2022-semeval},
qualitative evaluation is almost always presented in combination with these quantitative metrics. In this paper, we evaluate generated definitions with automatic metrics and by collecting human judgements.\looseness-1

\subsection{Semantic Change Detection}
Words in natural language change their meaning over time; these diachronic processes are of interest to both linguists and NLP practitioners.  Lexical semantic change detection (LSCD) is nowadays a well represented NLP task, with workshops \cite{lchange-2022-approaches} and several shared tasks 
\cite[e.g.,][]{schlechtweg-etal-2020-semeval,kurtyigit-etal-2021-lexical}. LSCD is usually cast either as binary classification (whether the target word changed its meaning or not) or as a ranking task (ordering target words according to the degree of their change). To evaluate existing approaches, manually annotated datasets are used: so-called DWUGs are described below in §\ref{sec:data}.\looseness-1


An important issue with current LSCD methods is that they rarely describe change in terms of \textit{word senses}, which are extremely important for linguists to understand diachronic meaning trajectories. Instead, systems provide (and are evaluated by) per-word numerical `change scores' which are hardly interpretable; at best, a binary `sense gain' or `sense loss' classification is used. Even approaches that do operate on the level of senses \cite[e.g.,][]{mitra2015automatic,homskiy-arefyev-2022-black} do not label 
them in a linguistically meaningful way,
making it difficult to understand the relations between the resulting `anonymous' types of word usage.

\section{Data}
\label{sec:data}
\subsection{Datasets of Definitions}
\label{sec:data-definition}
To train an NLG system that produces definitions (§\ref{sec:method}),
we use three datasets 
containing a human-written definition for each lexicographic sense of a target word, paired with a usage example.
\begin{table}[b]
\centering
\resizebox{\columnwidth}{!}{%
\begin{tabular}{@{}lrrccc@{}}
\toprule
\textbf{Dataset} & \textbf{Entries} & \textbf{Lemmas} & \textbf{Ratio} & \textbf{Usage length} & \textbf{Definition length} \\
\midrule
\textbf{WordNet} & 15,657  & 8,938  & 1.75 & 4.80 $\pm$ 3.43   & 6.64 $\pm$ 3.77  \\
\textbf{Oxford}  & 122,318 & 36,767 & 3.33 & 16.73 $\pm$ 9.53  & 11.01 $\pm$ 6.96 \\
\textbf{CoDWoE}  & 63,596  & 36,068 & 2.44 & 24.04 $\pm$ 21.05 & 11.78 $\pm$ 8.03\\
\bottomrule
\end{tabular}%
}
\caption{Main statistics of the datasets of definitions. Ratio is the \textit{sense-lemma} ratio: the number of entries over the number of lemmas.}
\label{tab:def-data-stats}
\end{table}
The \textbf{WordNet} dataset is a collection of word definitions and word usages extracted by \citet{ishiwatari-etal-2019-learning} from the WordNet lexical database~\cite{miller1995wordnet}.
The \textbf{Oxford} dataset (also known as CHA in prior work) consists of definitions and usage examples collected by \citet{gadetsky-etal-2018-conditional} from the Oxford Dictionary. 
Definitions are written by experts and usage examples are in British English. 
The \textbf{CoDWoE} dataset~\cite{mickus-etal-2022-semeval} is based on definitions and examples extracted from Wiktionary.\footnote{\url{https://www.wiktionary.org}} It is a multilingual corpus, of which we use the English portion. 
Table~\ref{tab:def-data-stats} reports the main statistics of these datasets. 
Further statistics, e.g., on the size of the different splits, are provided by \citet{huang-etal-2021-definition} as well as in Appendix~\ref{sec:appendix-data-stats}.\footnote{
	Note that in theory, a definition dataset could be also be extracted from the SemCor corpus \cite{miller-etal-1993-semantic}. However, we do not anticipate it will contribute much to training or evaluation since SemCor does not contain any new definitions with respect to WordNet: only more examples for the same word-definition pairs. This can be investigated in future work.
}




\begin{figure}
    \centering
    \includegraphics[width=0.8\linewidth]{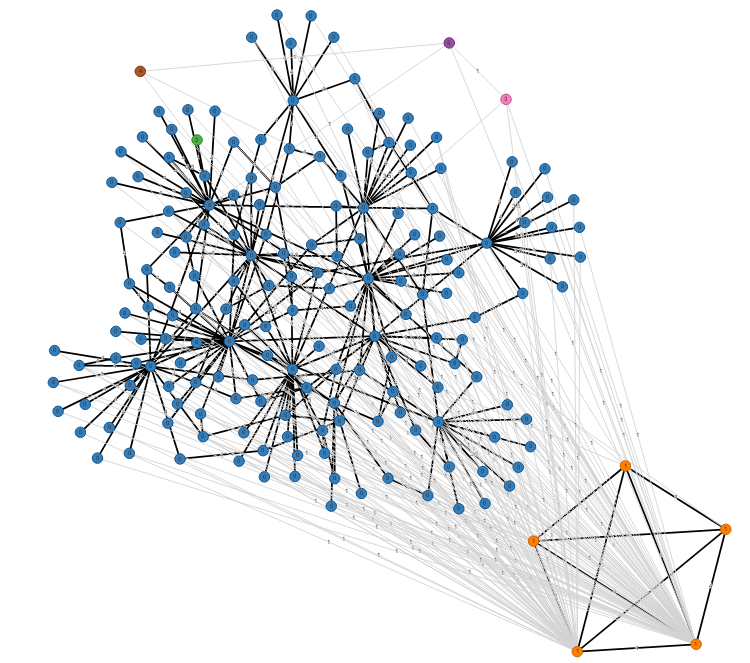}
    \caption{Diachronic word usage graph for the English word \word{lass} \cite{schlechtweg-etal-2021-dwug}.}
    \label{fig:lass_dwug}
\end{figure}

\subsection{Diachronic Word Usage Graphs}
\label{sec:data-dwugs}
We showcase interpretable word usage (§\ref{sec:pairwise}) and sense representations (§\ref{sec:labels} and \ref{sec:explain}) using a dataset where target lemmas are represented with diachronic word usage graphs~\cite[DWUGs,][]{schlechtweg-etal-2021-dwug}.
A DWUG is a weighted, undirected graph, where nodes represent target usages (word occurrences within a sentence or discourse context) and edge weights represent the semantic proximity of a pair of usages. 
DWUGs are the result of a multi-round incremental human annotation process, with annotators asked to judge the semantic relatedness of pairs of word usages on a 4-point scale. Based on these pairwise relatedness judgements, word usages are then grouped into usage clusters (a data-driven approximation of \textit{word senses}) using a variation of correlation clustering \cite{bansal-etal-2004-clustering,schlechtweg-etal-2020-semeval}.\looseness-1 

DWUGs are currently available in seven languages.\footnote{\url{https://www.ims.uni-stuttgart.de/en/research/resources/experiment-data/wugs/}} In this paper, we use the English graphs, which consist of usage sentences sampled from the Clean Corpus of Historical American English~\cite{davies2012expanding,alatrash-etal-2020-ccoha} and belonging to two time periods: 1810-1860 and 1960-2010. There are 46 usage graphs for English, corresponding to 40 nouns and 6 verbs annotated by a total of 9 annotators. Each target lemma has received on average 189 judgements, 2 for each usage pair.
Figure \ref{fig:lass_dwug} shows an example of a DWUG, with
colours denoting usage clusters (i.e., data-driven senses): the \textcolor{blue}{`blue'} and \textcolor{orange}{`orange'} clusters belong almost entirely to different time periods: a new sense of the word has emerged. We show how our approach helps explain such cases of semantic change in §\ref{sec:explain}.\looseness-1

\begin{table*}
\centering
\resizebox{0.9\textwidth}{!}{%
\begin{tabular}{llcccccc}
 & &
  \multicolumn{3}{c}{\textit{\textbf{WordNet}}} &
  \multicolumn{3}{c}{\textit{\textbf{Oxford}}} 
  \\ \toprule
  Model &
  Test &
  BLEU &
  ROUGE-L &
  \multicolumn{1}{c|}{BERT-F1} &
  BLEU &
  ROUGE-L &
  \multicolumn{1}{c}{BERT-F1} 
  \\ \midrule
\citet{huang-etal-2021-definition} & \textit{Unknown} &  32.72 & - & \multicolumn{1}{c|}{-} & \textbf{26.52} & - & \multicolumn{1}{c}{-} \\ 
   Flan-T5 XL & Zero-shot (task shift) &
   2.70 &  
   12.72 & 
  \multicolumn{1}{c|}{86.72} &
   2.88 &
   16.20 &
  \multicolumn{1}{c}{86.52}
   \\
   Flan-T5 XL & In-distribution &
   11.49 &
   28.96 & 
  \multicolumn{1}{c|}{88.90} &
   16.61 &
   36.27 &
  \multicolumn{1}{c}{89.40} 
   \\
  Flan-T5 XL & Hard domain shift &
   29.55 &
   48.17 &
  \multicolumn{1}{c|}{91.39} &
   8.37 &
   25.06 &
  \multicolumn{1}{c}{87.56} 
   \\
   Flan-T5 XL & Soft domain shift &
   \textbf{32.81} &
   \textbf{52.21} &
  \multicolumn{1}{c|}{\textbf{92.16}} &
  18.69 &
  \textbf{38.72} &
  \multicolumn{1}{c}{\textbf{89.75}} 
   \\ \bottomrule
\end{tabular}%
}
\caption{Results of the definition generation experiments.}
\label{tab:def-gen-results}
\end{table*}

\section{Definition Generation}
\label{sec:method}

Our formulation of the \textit{definition generation} task is as follows: given a target word $w$ and an example usage $s$ (i.e., a sentence containing an occurrence of $w$), generate a natural language definition $d$ that is grammatical, fluent, and faithful to the meaning of the target word $w$ as used in the example usage $s$.
A \textit{definition generator} is a language process 
that maps words and example usages to natural language definitions. 
%
As a generator, we use Flan-T5 \cite{chung2022scaling}, a version of the T5 encoder-decoder Transformer~\cite{raffel2020exploring} fine-tuned on 1.8K tasks phrased as instructions and collected from almost 500 NLP datasets. Flan-T5 is not trained specifically on definition generation but thanks to its massive multi-task instruction fine-tuning, the model exhibits strong generalisation to unseen tasks. Therefore, we expect it to produce high-quality definitions. We extensively test three variants of Flan-T5 of different size and compare them to vanilla T5 models (Table~\ref{tab:pairwise} and Table~\ref{tab:def-gen-results-base}, Appendix~\ref{sec:appendix-other-models}); based on our results, we recommend using the largest fine-tuned Flan-T5 model whenever possible.

To obtain definitions from Flan-T5, we use
natural language prompts consisting of an example usage preceded or followed by a question or instruction. For example: `$s\ $ \texttt{What is the definition of $w$?}'.
The concatenated usage example and prompt are provided as input to Flan-T5, which conditionally generates definitions (Table~\ref{tab:draftee} shows an example instance).\footnote{
This is a simpler workflow in comparison to prior work \cite{bevilacqua-etal-2020-generationary,almeman-espinosa-anke-2022-putting} where inputs are encoded as `target word - context' pairs.} We choose greedy search with target word filtering as a simple, parameter-free decoding strategy. Stochastic decoding algorithms can be investigated in future work.\looseness-1

\paragraph{Prompt selection}
In preliminary experiments, we used the pre-trained Flan-T5 Base model (250M parameters) to select a definition generation prompt among 8 alternative verbalisations.
Appending the question \textit{`What is the definition of $w$?'} to the usage example consistently yielded the best scores.\footnote{
Further details in Appendix \ref{sec:app-prompt-selection}.}
We use this prompt for all further experiments.

\subsection{Evaluating Generated Definitions}
\label{sec:method-def-eval}
Before using its definitions to construct an interpretable semantic space---the main goal of this paper---we perform a series of experiments to validate Flan-T5 as a definition generator. 
We use the target lemmas and usage examples from the corpora of definitions presented in §\ref{sec:data}, conditionally generate definitions with Flan-T5, and then compare them to the gold definitions in the corpora 
using reference-based NLG evaluation metrics. We report SacreBLEU and ROUGE-L, which measure surface form overlap, as well as BERT-F1, which is sensitive to the reference and candidate's semantics. 
As mentioned in §\ref{subsec:background-def-generation}, reference-based metrics are not flawless,
yet designing and validating a reference-free metric for the definition generation task is beyond the scope of this paper.
We will later resort to correlations with human judgements and expert human evaluation to assess the quality of generated definitions.\looseness-1
 
We evaluate the Flan-T5 XL (3B parameters) in three generalisation tests: 
1)~in distribution, 2)~hard domain shift, and 3)~soft domain shift.\footnote{
Tests defined following the GenBench generalisation taxonomy~\citep{hupkes2022state}.
We also include a fourth setup, \textit{zero shot (task shift)}, where we directly evaluate the pretrained Flan-T5 XL. Results (including other models) are presented in Appendix~\ref{sec:appendix-zero-shot}-\ref{sec:appendix-other-models}, and an evaluation card in Appendix~\ref{sec:appendix-eval-cards}.
}
We use these tests to choose a model to be deployed in further experiments. For reference, we report the BLEU score of the definition generator by \citet{huang-etal-2021-definition}; ROUGE-L and BERT-F1 are not reported in their paper.

\paragraph{In distribution} We fine-tune Flan-T5 XL on one corpus of definitions at a time, and test it on a held-out set from that same corpus (except CoDWoE which does not provide train-test split). The quality of the definitions increases substantially with fine-tuning, in terms of both their lexical and semantic overlap with gold definitions (Table~\ref{tab:def-gen-results}). We find significantly higher scores on Oxford,
which may be due to the larger size of its training split and to the quality of the WordNet examples, which sometimes are not sufficiently informative~\citep{almeman-espinosa-anke-2022-putting}.\looseness-1

\paragraph{Hard domain shift} We fine-tune Flan-T5 XL on WordNet and test it on Oxford, and vice versa. These tests allow us to assess the model's sensitivity to the peculiarities of the training dataset. A model that has properly learned to generate definitions should be robust to this kind of domain shift. 
The quality of the definitions of Oxford lemmas generated with the model fine-tuned on WordNet (see the Oxford column in Table~\ref{tab:def-gen-results}) is lower than for the model fine-tuned on Oxford itself (same column, see row `In-distribution'). Instead, for out-of-domain WordNet definitions, all metrics surprisingly indicate higher quality than for in-distribution tests (WordNet column). Taken together, our results so far suggest that the quality of a fine-tuned model depends more on the amount (and, possibly, on the quality) of the training data than on whether the test data is drawn from the same dataset. \looseness-1

\paragraph{Soft domain shift} We finally fine-tune Flan-T5 XL on a collection of all three definition datasets: WordNet, Oxford, and CoDWoE. Our previous results hint towards the model's preference for more training examples, so we expect this setup to achieve the highest scores regardless of the soft shift between training and test data. Indeed, on WordNet, our fine-tuned model marginally surpasses the state-of-the-art upper bound in terms of BLEU score (Table~\ref{tab:def-gen-results}), and it achieves the highest scores on the other metrics. Oxford definitions generated with this model are instead still below Huang et al.'s upper bound; this may be due to Oxford being generally more difficult to model than WordNet, perhaps because of longer definitions and usages (see Figures~\ref{fig:wordnet}-\ref{fig:oxford} in Appendix~\ref{sec:appendix-data-stats}).
We consider 
the observed model performance sufficient 
for the purposes of our experiments, in particular in view of the higher efficiency of fine-tuned Flan-T5 with respect to the three-module system of \citet{huang-etal-2021-definition}. We therefore use this model throughout the rest of our study.\looseness-1

The Flan-T5 models fine-tuned for definition generation are publicly available through the Hugging Face model hub.\footnote{
    Model names: \href{https://huggingface.co/ltg/flan-t5-definition-en-base}{ltg/flan-t5-definition-en-base}, \href{https://huggingface.co/ltg/flan-t5-definition-en-large}{ltg/flan-t5-definition-en-large}, \href{https://huggingface.co/ltg/flan-t5-definition-en-xl}{ltg/flan-t5-definition-en-xl}.
}


\section{Definitions are Interpretable Word Representations}
\label{sec:pairwise}
%
We propose considering the abstract meaning space of definitions as a representational space for lexical meaning. Definitions fulfil important general desiderata for word representations: they are human-interpretable and they can be used for quantitative comparisons between word usages (i.e., by judging the distance between pairs of definition strings). We put the \textit{definition space} to test by applying it to the task of semantic change analysis, which requires capturing word meaning at a fine-grained level, distinguishing word senses based on usage contexts.
We use our fine-tuned Flan-T5 models (XL and other sizes)
to generate definitions for all usages of the 46 target words annotated in the English DWUGs (ca.\ 200 usages per word; see §\ref{sec:data-dwugs}).\footnote{The training datasets used in §\ref{sec:method} contain nouns, verbs, adjectives and adverbs. The English DWUGs contain only nouns and verbs.}
These definitions (an example is provided in Table~\ref{tab:draftee}) specify a diachronic semantic space.\looseness-1

\subsection{Correlation with Human Judgements}
\label{sec:correlation}

\begin{table}
\centering
\resizebox{\columnwidth}{!}{%
\begin{tabular}{lccc}
\toprule
\textbf{Method}    & \textbf{Cosine}   & \textbf{SacreBLEU} & \textbf{METEOR}\\
\midrule
Token embeddings   & 0.141    &  - & - \\
Sentence embeddings & 0.114    &  - & - \\
\midrule
\multicolumn{4}{c}{\textbf{Generated definitions}} \\
\midrule
Flan-T5 XL Zero-shot       & 0.188    & 0.041    &  0.083 \\
Flan-T5 XXL Zero-shot       & 0.206    & 0.045     & 0.092 \\
Flan-T5 base FT      & 0.221    & 0.078    & 0.077 \\
Flan-T5 XL FT        & \textbf{0.264}    & \textbf{0.108}   &  \textbf{0.117} \\
\bottomrule
\end{tabular}%
}
\caption{Correlations with pairwise similarity judgements by humans. `FT' stands for `fine-tuned model'.}
\label{tab:pairwise}
\end{table}

We construct word usage graphs for each lemma in the English DWUGs: we take usages as nodes and assign weights to edges by measuring pairwise similarity between usage-dependent definitions.
We compute the similarity between pairs of definitions using two overlap-based metrics, SacreBLEU and METEOR, as well as the cosine similarity between sentence-embedded definitions. 
We then compare our graphs against the gold DWUGs, where edges between usage pairs are weighted with human judgements of semantic similarity, by computing the Spearman's correlation between human similarity judgements and similarity scores obtained for pairs of generated definitions. We compare our results to DWUGs constructed based on two additional types of usage-based representations: \textit{sentence} embeddings obtained directly for usage examples, and contextualised \textit{token} embeddings. Sentence embeddings (for both definitions and usage examples) are SBERT representations \cite{reimers-gurevych-2019-sentence} extracted with mean-pooling from the last layer of a DistilRoBERTa LM fine-tuned for semantic similarity comparisons.\footnote{
DistilRoBERTa (\texttt{sentence-transformers/all-} \texttt{distilRoBERTa-v1}) is the second best model as reported in the official S-BERT documentation at the time of publication (\url{https://www.sbert.net/docs/pretrained_models.html}). For a negligible accuracy reduction, it captures longer context sizes and is ca.~50\% smaller and faster than the model that ranks first.}  
For tokens, we extract the last-layer representations of a RoBERTa-large model \cite{liu2019RoBERTa} which correspond to subtokens of the target word \cite[following][]{giulianelli-etal-2020-analysing} and use mean-pooling to obtain a single vector. 
While we report string-overlap similarities for definitions, these are not defined for numerical vectors, and thus similarities for example sentences and tokens are obtained with cosine only.\looseness-1

Pairwise similarities between definitions approximate human similarity judgements far better than similarities between example sentence and word embeddings (Table~\ref{tab:pairwise}). 
This indicates that definitions are a more accurate approximation of contextualised lexical meaning. 
The results also show that similarity between definitions is best captured by their embeddings, rather than by overlap-based metrics like SacreBLEU and METEOR.

\subsection{Definition Embedding Space}

\begin{figure}
    \centering
    \includegraphics[trim={7cm 2cm 7cm 1cm},clip, width=\linewidth]{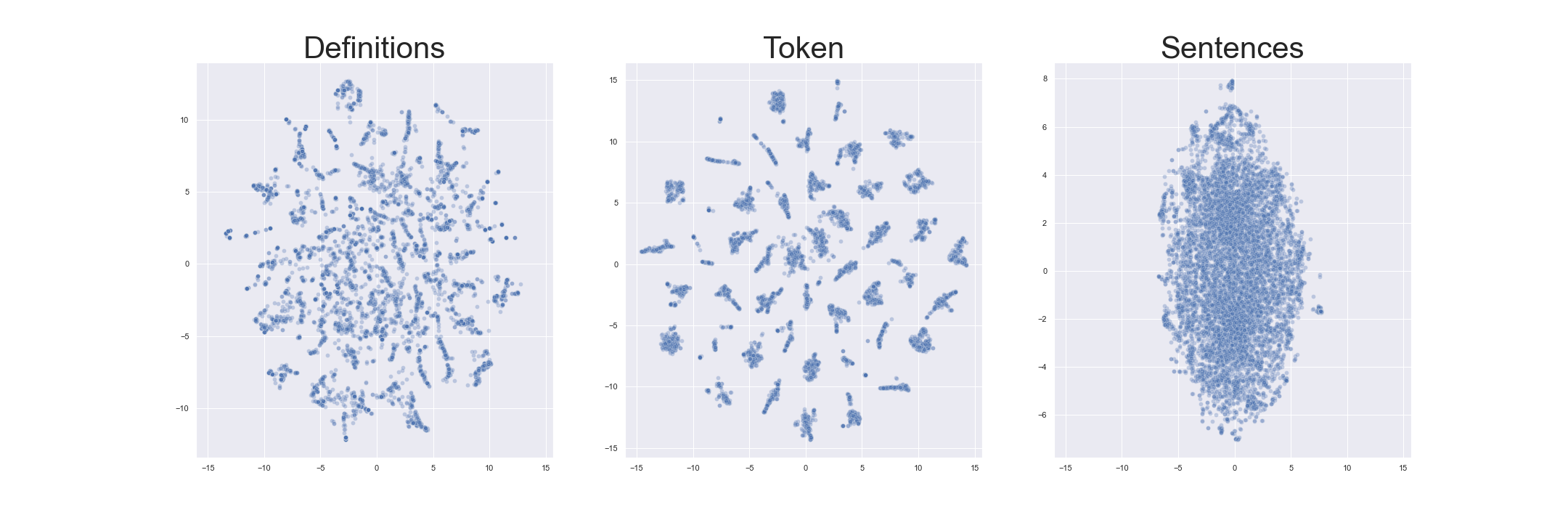}
    \caption{T-SNE projection of each embedding space, DistilRoBERTa model.}
    \label{fig:T-SNE-distil}
\end{figure}

We now examine the \textit{definition embedding space} (the high-dimensional semantic space defined by sentence-embedded definitions), to identify properties that make it more expressive than usage-based spaces.
Figure \ref{fig:T-SNE-distil} shows the T-SNE projections of the DistilRoBERTa embeddings of all lemmas in the English DWUGs, for the three types of  representation presented earlier: generated definitions, tokens, and example sentences.\footnote{T-SNE projections for RoBERTa-large are in Appendix~\ref{sec:appendix-embedding-spaces}.
}
The definition spaces 
exhibit characteristics that are more similar to a \textit{token} embedding space than an example \textit{sentence} embedding space, with definitions of the same lemma represented by relatively close-knit clusters of definition embeddings. 
This suggests that definition embeddings, as expected, represent the meaning of a word in context (similar to token embeddings), rather than the meaning of the whole usage example sentence in which the target word occurs.\looseness-1

For each target word, we also measure (i) the variability in each embedding space and (ii) the inter-cluster and intra-cluster dispersion~\cite{calinski1974dendrite} 
obtained when clustering each space using $k$-means.
This allows us to quantitatively appreciate properties exhibited by data-driven usage clusters that are obtained from different representation types.
To cluster the embedding spaces, we experiment with values of $k \in [2, 25]$, and select the $k$ which maximises the Silhouette score. Our results are summarised in Table~\ref{tab:Space_averages}.
%
While, on average, token spaces exhibit higher inter-cluster dispersion (indicating better cluster separation), the clusters in the definition spaces have on average the lowest intra-cluster dispersion, indicating that they are more cohesive than the clusters in the token and example sentence spaces.  
These findings persist for the gold clusters determined by the English DWUGs (Table~\ref{tab:DWUG_averages}, Appendix~\ref{sec:appendix-embedding-spaces}).

In sum, this analysis shows that definition embedding spaces are generally suitable to distinguish different types of word usage. In the next section, we will show how they can indeed be used to characterise word senses.

\begin{table}
    \resizebox{\columnwidth}{!}{%
    \begin{tabular}{llrrrrrrrr}
    \toprule
    \textbf{Model}  & \textbf{Representation} & \textbf{Variance} & \textbf{Std} & \textbf{$K$}&  \textbf{Silh.} $\uparrow$ & \textbf{Sep.} $\uparrow$ & \textbf{Coh.} $\downarrow$ & \textbf{Ratio} $\uparrow$\\
    \midrule
       &  Sentence             &  0.014    &   0.117   &  2.0  &  0.111  &  0.285 &   0.012 &  23.2 \\
    RoBERTa-large                & Token                 &  0.034    &   0.183   &  3.8  &  0.173  &  \textbf{0.868} &   0.027  &  \textbf{32.4}  \\
                    & Definitions           &  0.006    &   0.080   &  20.6 &  \textbf{0.335}  &  0.057 &   \textbf{0.003} &  22.3\\ \midrule
     & Sentence                &  0.597    &   0.772   &   2.1 &  0.046  &  4.907 &   0.578 &   8.5\\
    DistilRoBERTa                & Token                 &  0.477    &   0.687   &   2.5 &  0.121  &  \textbf{8.599} &   0.427  &  20.1\\
                    & Definitions           &  0.509    &  0.756    &  19.7 &  \textbf{0.355}  &  5.559 &    \textbf{0.228} &  \textbf{24.4}\\
    \bottomrule
    \end{tabular}
    }
 \caption{Variance, standard deviation, optimal $K$, silhouette score, separation score, cohesion score, and the separation-cohesion ratio  for each embedding space; average over all target words.}
\label{tab:Space_averages}
\end{table}

\section{Labelling Word Senses With Definitions}
\label{sec:labels}

For generated definitions to be useful in practice, they need to be able to distinguish word senses.
%
For example (ignoring diachronic differences and singleton clusters), there are three main senses of the word \word{word} in its DWUG, which we manually label as: (1) \sense{words of language}, (2) \sense{a rumour}, and (3)~\sense{an oath}. 
%
Manual inspection of the generated definitions indicates that they
are indeed sense-aware:\looseness-1
\begin{enumerate}
\itemsep=0em
    \item \textit{`A communication, a message', `The text of a book, play, movie'}, etc.
    \item \textit{`Information passed on, usually by one person to another', `communication by spoken or written communication'}, etc.
    \item \textit{`An oath', `a pronouncement'}, etc.
\end{enumerate}


%
But let's again put ourselves in the shoes of a historical linguist. 
Sense clusters are now impractically represented with multitudes of contextualised definitions. Cluster (1) for \word{word}, e.g., features $190$ usages, and
one must read through all of them 
(otherwise there will be a chance of missing something) and generalise -- all 
to formulate a definition that covers the whole sense cluster (a \textit{sense label}).
We now show how DWUGs can be automatically augmented with generated sense labels, vastly improving their usability.\looseness-1

\paragraph{Selecting sense labels}
From $n$ definitions, generated for $n$ word usages belonging to the same DWUG cluster, we use the most prototypical one as the \textit{sense label}---with the aim of reflecting the meaning of the majority of usages in the cluster.
We represent all definitions with their sentence embeddings (cf.~\S\ref{sec:correlation}) and select as prototypical the definition whose embedding is most similar to the average of all embeddings in the cluster.
Clusters with less than 3 usages are ignored as, for these, prototypicality is ill-defined. 
As a sanity check, these are the sense labels obtained by this method for the DWUG clusters of \word{word}; they correspond well to the sense descriptions provided earlier.
\begin{enumerate}
\itemsep=0em 
    \item \sense{A single spoken or written utterance}
    \item \sense{Information; news; reports}
    \item \sense{A promise, vow or statement}
\end{enumerate}

%
We compare these sense labels to labels obtained by generating a definition for the most prototypical \textit{usage} (as judged by its token embedding), rather than taking the most prototypical \textit{definition}, and we evaluate both types of senses labels using human judgements. Examples of labels can be found in Appendix~\ref{sec:examples}.

\paragraph{Human evaluation}
Five human annotators (fluent English speakers)
were asked to evaluate the quality of sense labels for each cluster in the English DWUGs, 136 in total.
Each cluster was accompanied by the target word, two labels (from definitions and from usages) and five example usages randomly sampled from the DWUG. The annotators could select one of six judgements to indicate overall quality of the labels and their relative ranking. 
After a reconciliation round, the categorical judgements were aggregated via majority voting.  Krippendorff's $\alpha$ inter-rater agreement is $0.35$ on the original data and $0.45$ when the categories are reduced to four. Full guidelines and results are reported in Appendix~\ref{sec:guidelines}.\footnote{
	There exist no established procedures for the collection of human quality judgements of automatically generated word sense labels. The closest efforts we are aware of are those in \citet{noraset2017definition}, who ask annotators to rank definitions generated by two systems, providing as reference the gold dictionary definitions. In our case, (1)~generations are for word senses rather than lemmas, (2)~we are interested not only in rankings but also in judgements of `sufficient quality', (3)~dictionary definitions are not available for the DWUG senses; instead (4)~we provide annotators with usage examples, which are crucial for informed judgements of sense definitions. 
}

We find that our prototypicality-based sense labelling strategy is overall reliable.
Only for 15\% of the clusters, annotators indicate that neither of the labels is satisfactory
(Figure~\ref{fig:annot_quality}).  
When comparing definition-based and usage-based labels, the former were found to be better in 31\% of the cases, while the latter in only 7\% (in the rest of the cases, the two methods are judged as equal).
We also analysed how often the labels produced by each method were found to be acceptable. Definition-based labels were of sufficient quality in 80\% of the instances, while for usage-based labels this is only true for 68\% of the cases.\looseness-1 

In sum,  prototypical definitions reflect sense meanings better than definitions of prototypical usage examples. We believe this is because definitions are more abstract and robust to contextual noise (the same definition can be assigned to very different usages, if the underlying sense is similar). 
This approach takes the best of both worlds: the produced representations are data-driven, but at the same time they are human-readable and naturally explanatory. After all, `senses are abstractions from clusters of corpus citations' \cite{kilgarriff1997don}.
In the next section, we demonstrate how automatically generated definition-based sense labels can be used to explain semantic change observed in diachronic text corpora.


\section{Explaining Semantic Change with Sense Labels}
\label{sec:explain}

Word senses in DWUGs are collections of example usages and they are only labelled with numerical identifiers.
This does not allow users to easily grasp the meaning trajectories of the words they are interested in studying.
Using sense labels extracted from generated definitions, we can produce a fully human-readable \textit{sense dynamics map}---i.e., an automatically annotated version of a DWUG which displays synchronic and diachronic relations between senses (e.g,
senses transitioning one into another, splitting from another sense, or two senses merging into one). One can look at sense dynamics maps as reproducing the work of \newcite{mitra2015automatic} on the modern technological level and, importantly, with human-readable sense definitions.

Given a target word, its original DWUG, and its semi-automatic sense clusters, we start by assigning a definition label to each cluster, as described in §\ref{sec:labels}. Then, we divide each cluster into two sub-clusters, corresponding to time periods $1$ and $2$ (for example, sub-cluster $c_1^2$ contains all usages from cluster $1$ occurring in time period $2$).\footnote{  
	Note that the labels are still time-agnostic: that is, sub-clusters $c_1^1$ and $c_1^2$ have the same label. This is done for simplicity and because of data scarcity, but in the future we plan to experiment with time-dependent labels as well.
	We use two time periods as only two periods are available in Schlechtweg et al.'s English DWUGs \shortcite{schlechtweg-etal-2021-dwug}, but the same procedure can be executed on multi-period datasets.
}
%
We compute pairwise cosine similarities between the sentence embeddings of the labels (their `definition embeddings'),
thereby producing a fully connected graph where nodes are sub-clusters and edges are weighted with sense label similarities. Most edges have very low weight,
but some sub-cluster pairs are unusually similar, hinting at a possible relation between the corresponding senses.
\begin{figure*}
    \centering
    \includegraphics[width=0.4\linewidth]{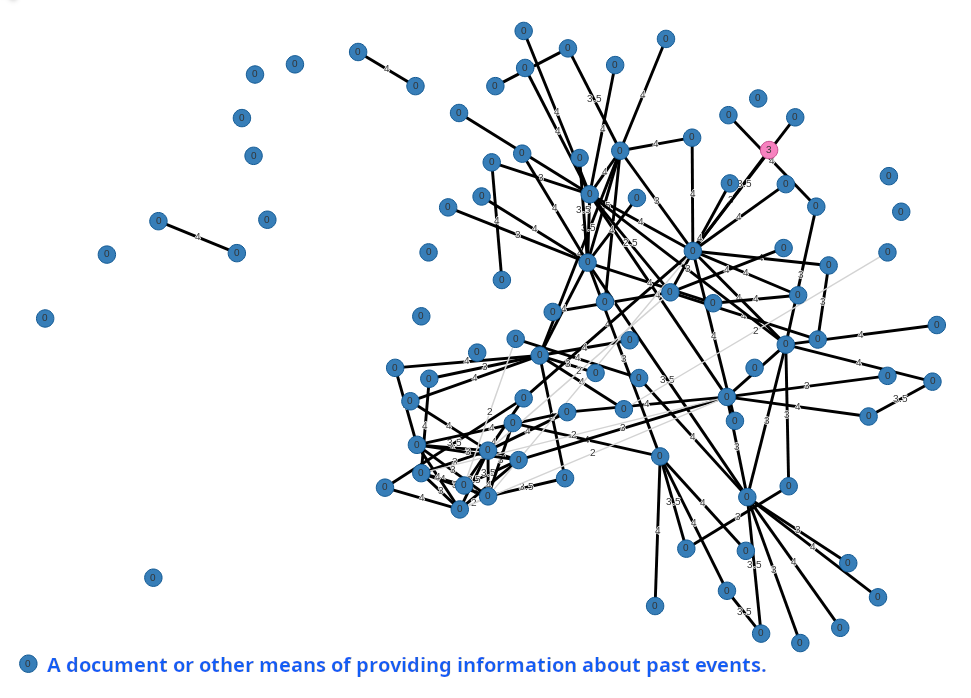}
     \includegraphics[width=0.4\linewidth]{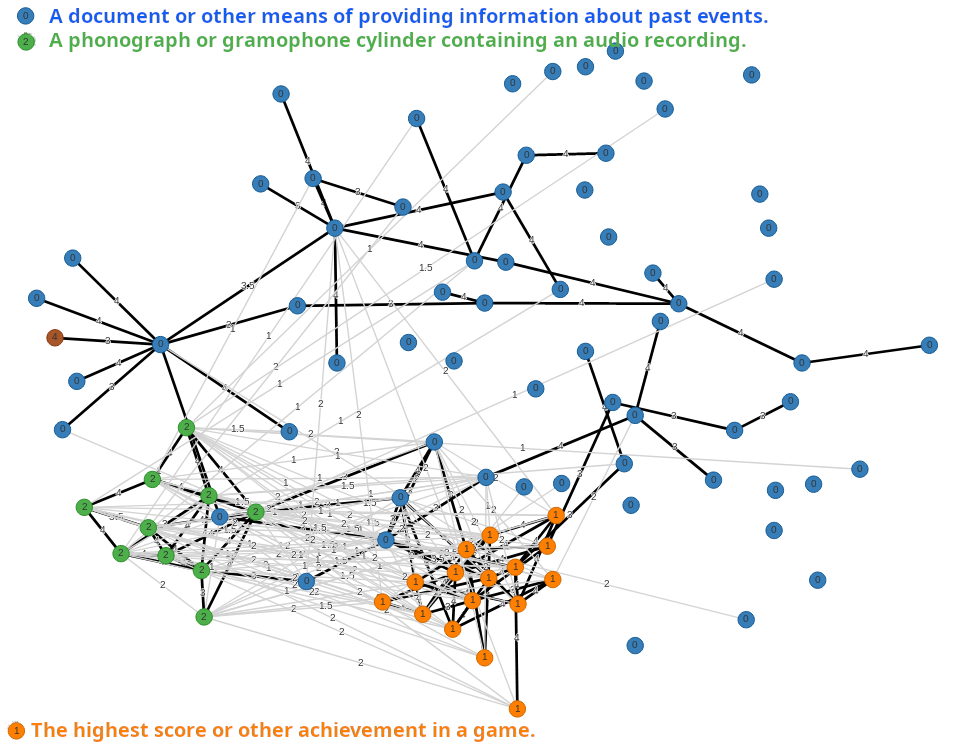}
    \caption{Diachronic word usage graphs for \word{record} \cite{schlechtweg-etal-2021-dwug} with sense definitions generated using our proposed procedure (\S\ref{sec:labels}). Left: time period 1 (1810-1860); right: time period 2 (1960-2010). Colours correspond to data-driven senses, as annotated in the original DWUGs.}
    \label{fig:record}
\end{figure*}
We detect these outlier pairs by inspecting the distribution of pairwise similarities for values with $z$-score higher than~$1$ (similarities more than $1$ standard deviation away from the mean similarity).
Sub-cluster pairs connected with such edges form a \textit{sense dynamics map}.
\looseness-1 

As an example, 
the noun \word{record} has only one sense in time period 1 
but it acquires two new senses in time period 2  (Figure~\ref{fig:record}; as before, we ignore clusters with less than 3 usages). 
The sense clusters defined by the DWUG are anonymous collection of usages,
but with the assigned sense labels (also shown in Figure~\ref{fig:record})
they can be turned into a proto-explanation of the observed semantic shift:
\begin{itemize}
    \item A novel sense $2$ of \word{record} in time period $2$ (\sense{A phonograph or gramophone cylinder containing an audio recording.}) is probably an offshoot of a stable sense $0$ present in both time periods (\sense{A document or other means of providing information about past events.}).
\end{itemize}
It becomes now clear that sense $2$ stems from the older general sense $0$ of \word{record}---arguably representing a case of narrowing \cite{bloomfield}---while the second new sense ($1$: \sense{the highest score or other achievement in the game}) is not related to the others and is thus independent.

%

Sense dynamics maps can also help in tracing potentially incorrect or inconsistent clustering in DWUGs. For instance, if different sense clusters are assigned identical definition labels, then it is likely that both clusters correspond to the same sense and that the clustering is thus erroneous.
Using our automatically produced sense dynamics maps, DWUGs can be improved and enriched (semi-)automatically.

An interesting 
case is \word{ball} (see Appendix~\ref{sec:appendix-maps} for another example regarding the word \word{chef}). Although none of its sense labels are identical, its sense cluster $c_0$ is very close to cluster $c_2$ (similarity of $0.70$), while $c_2$ is close to $c_3$ (similarity of $0.53$); all three senses persist throughout both time periods, with sense $3$ declining in frequency. The generated definitions for the \word{ball} clusters are: $0$: \sense{A sphere or other object used as the object of a hit} (the largest cluster), $2$: \sense{A round solid projectile, such as is used in shooting}, and $3$: \sense{A bullet}. 
%
This case
demonstrates that similarity relations are not transitive: the similarity between $c_0$ and $c_3$ is only $0.50$, below our outlier threshold value. 
This is in part caused by inconsistent DWUG clustering: while the majority of usages in $c_2^1$ are about firearm projectiles, $c_2^2$ contains mentions of golf balls and ball point pens. This shifts sense $2$ from \sense{bullet} to \sense{round solid projectile}, making it closer to sense $0$ (general spheres) than it should be. Ideally, all the \sense{bullet} usages from $c_2$ should have ended up in $c_3$, with the rest joining the general sense $0$.\looseness-1 

Besides suggesting fixes to the DWUG clustering, the observed non-transitivity also describes a potential (not necessarily diachronic) meaning trajectory of \word{ball}: from any spherical object, to spherical objects used as projectiles, and then to any projectiles (like bullets), independent of their form. Our generated sense labels and their similarities help users analyse this phenomenon in a considerably faster and easier way than by manually inspecting all examples for these senses.

\section{Conclusion and Future Work}
\label{sec:conclusion}

In this paper, we propose to consider automatically generated contextualised word definitions as a type of lexical representation, similar to traditional word embeddings. 
While generated definitions have been already shown to be effective for word sense disambiguation \cite{bevilacqua-etal-2020-generationary}, our study puts this into a broader perspective and demonstrates that 
modern language models like Flan-T5 \cite{chung2022scaling} are sufficiently mature to produce robust and accurate definitions in a simple prompting setup.
The generated definitions outperform traditional token embeddings in word-in-context similarity judgements while being naturally interpretable.\looseness-1

We apply definition-based lexical representations to semantic change analysis and show that our approach can be used to trace word sense dynamics over time. 
Operating in the space of human-readable definitions makes such analyses much more interesting and actionable for linguists and lexicographers---who look for explanations, not numbers.
At the same time, we believe the `definitions as representations' paradigm can also be used for other NLP tasks in the area of lexical semantics, such as word sense induction, idiom detection, and metaphor interpretation.

Our experiments with diachronic sense modelling are still preliminary and mostly qualitative. It is important to evaluate systematically how well our predictions correspond to the judgements of (expert) humans. Once further evidence is gathered, other promising applications include tracing cases of semantic narrowing or widening over time \cite{bloomfield} by analysing the variability of contextualised definitions in different time periods and by making cluster labels time-dependent. Both directions will require extensive human annotation, and we leave them for future work.



\section*{Limitations}
Data in this work is limited to the English diachronic word usage graphs (DWUGs). Our methods themselves are language-agnostic and we do not anticipate serious problems with adapting them to DWUGs in other languages (which already exist). At the same time, although Flan-T5 is a multilingual LM, we did not thoroughly evaluate its ability to generate definitions in languages other than English. Again, definition datasets in other languages do exist and technically it is trivial to fine-tune Flan-T5 on some or all of them.

Generated definitions and mappings between definitions and word senses can contain all sorts of biases and stereotypes, stemming from the underlying language model. Filtering inappropriate character strings from the definitions can only help as much, and further research is needed to estimate possible threats.

In our experiments with Flan-T5, the aim was to investigate the principal possibility of using this LM for definition modelling. Although we did evaluate several different Flan-T5 variants, we leave it for the future work to investigate the impact of model size and other experimental variables (such as decoding algorithms).

The cases shown in §\ref{sec:explain} are hand-picked examples, demonstrating the potential of using generated definitions for explainable semantic change detection and improving LSCD datasets. In the future, we plan to conduct a more rigorous evaluation of different ways to build sense dynamics map.

\section*{Acknowledgements}
This project has received funding from the European Research Council (ERC) under the European Union's Horizon 2020 research and innovation programme (grant agreement No.\ 819455).
The computations were performed on resources provided through Sigma2---the national research infrastructure provider for High-Performance Computing and large-scale data storage in Norway.

\bibliography{anthology,custom}
\bibliographystyle{acl_natbib}

\appendix
\section*{Appendix}

\section{Preliminary Analysis of Usage Examples}
\label{sec:appendix-data-stats}
\begin{figure}
    \centering
    \includegraphics[width=0.2\textwidth]{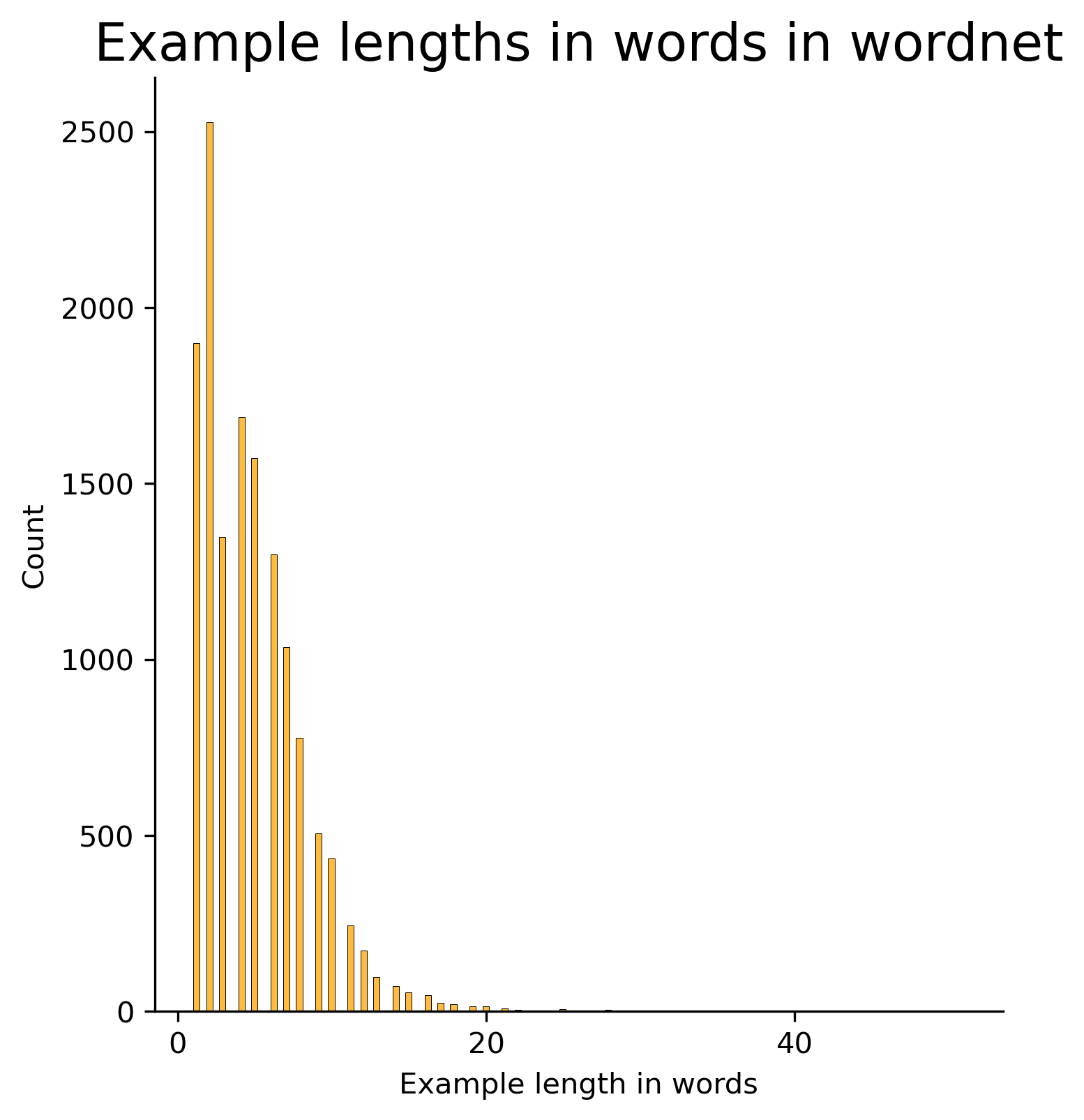}
    \includegraphics[width=0.2\textwidth]{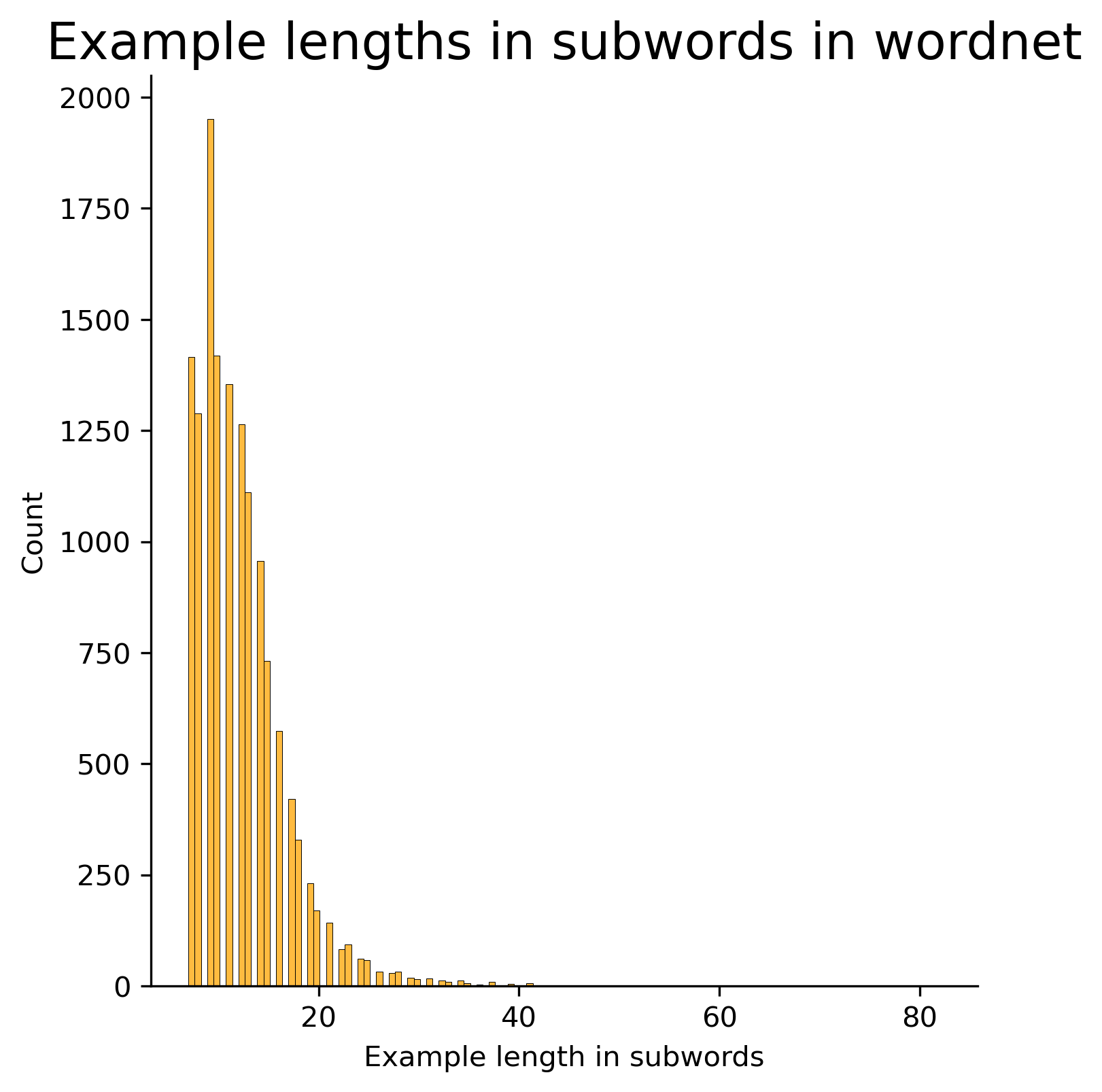}
    \caption{Length distribution of usage examples in WordNet.}
    \label{fig:wordnet}
\end{figure}

\begin{figure}
    \centering
    \includegraphics[width=0.2\textwidth]{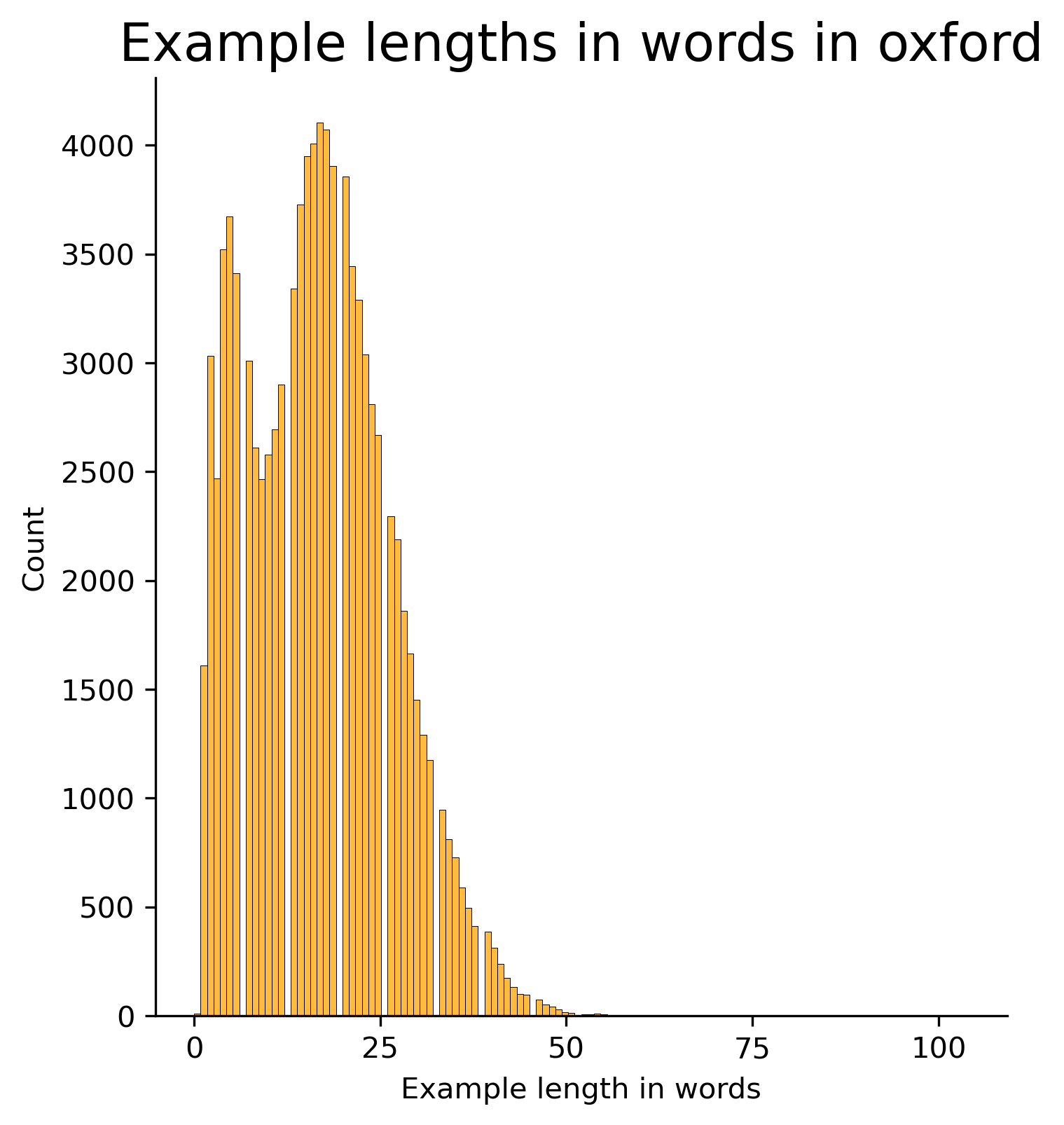}    
    \includegraphics[width=0.2\textwidth]{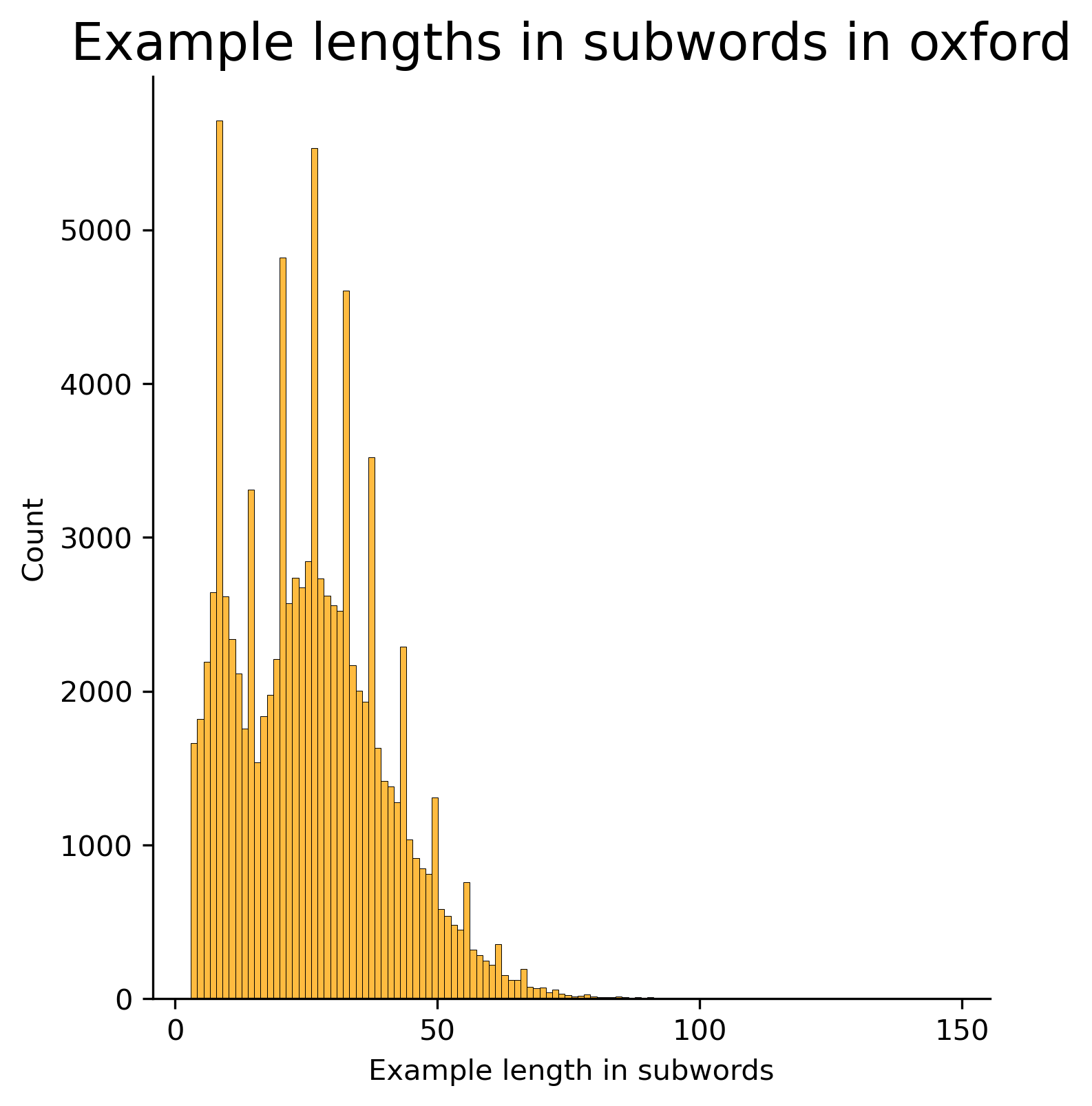}
    \caption{Length distribution of usage examples in Oxford.}
    \label{fig:oxford}
\end{figure}

\begin{figure}
    \centering
    \includegraphics[width=0.2\textwidth]{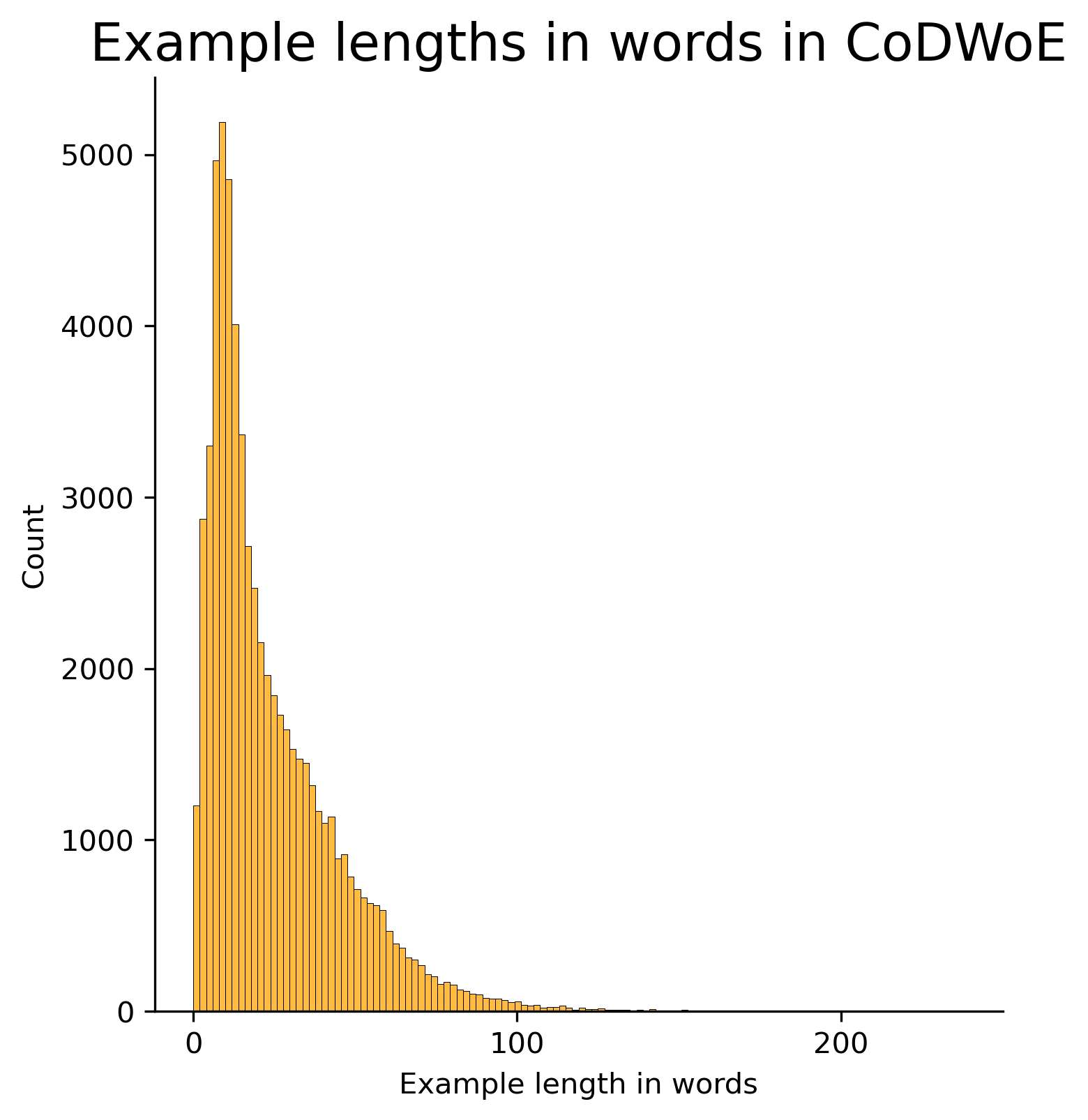}
    \includegraphics[width=0.2\textwidth]{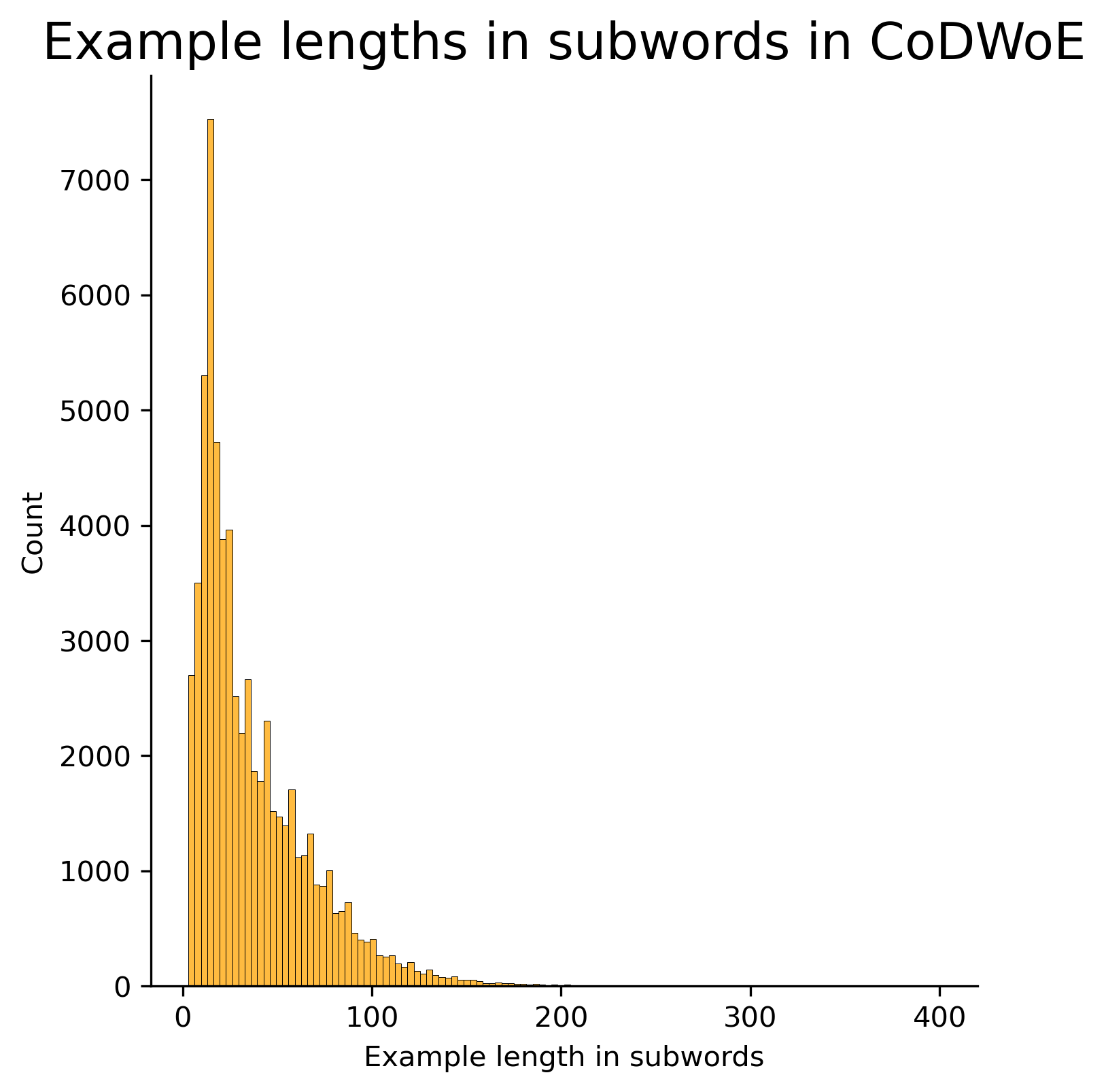}
    \caption{Length distribution of usage examples in CoDWoE.}
    \label{fig:codwoe}
\end{figure}

\begin{figure}
    \centering
    \includegraphics[width=0.2\textwidth]{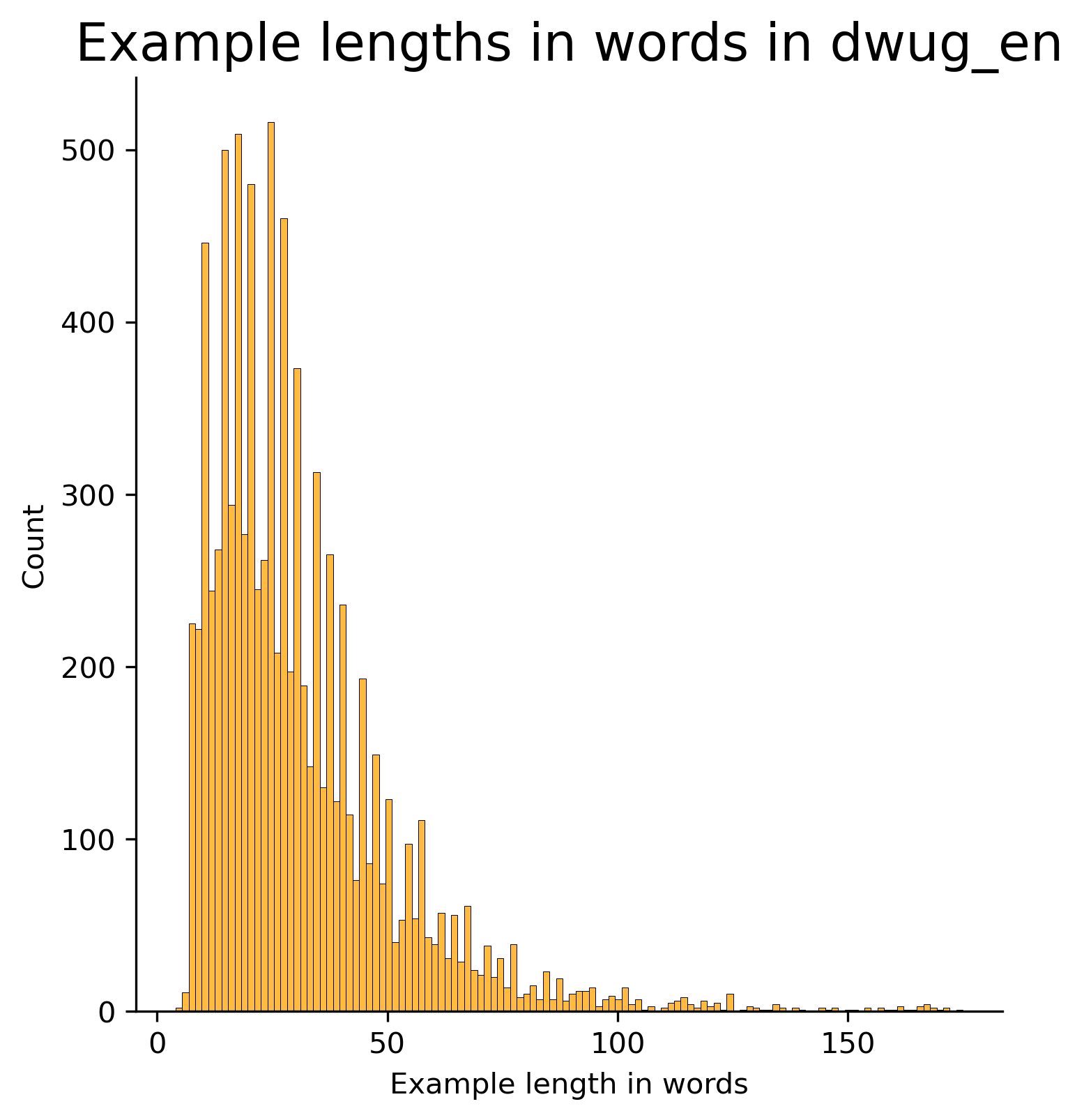}
    \includegraphics[width=0.2\textwidth]{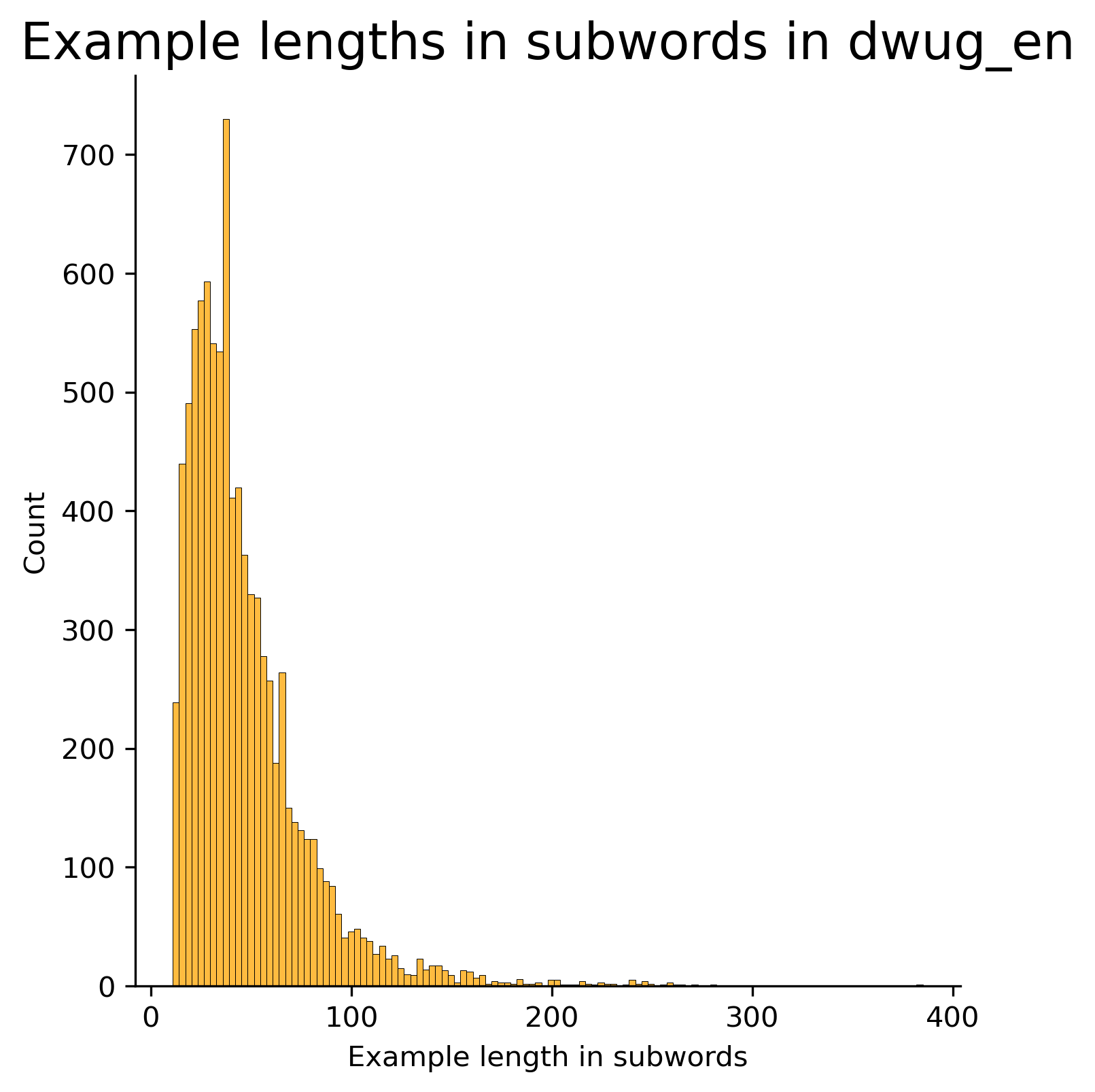}
    \caption{Length distribution of usage examples in the English DWUGs.}
    \label{fig:dwug_en}
\end{figure}

In Section~\ref{sec:data-definition} of the main paper, we present three corpora of human-written definitions and report their main statistics in Table~\ref{tab:def-data-stats}, including mean and standard deviation of usage example length. Because the length of usage examples has been shown to affect the quality of generated definitions~\cite{almeman-espinosa-anke-2022-putting}, in a preliminary analysis, we compare the length distributions of usage examples in the corpora of definitions as well as in the English DWUGs~\cite{schlechtweg-etal-2021-dwug}.
Figures~\ref{fig:wordnet}-\ref{fig:dwug_en} show the length distributions of the four datasets.
We also measure the correlation between definition quality (BertScore, BLEU, NIST) and (i) the length of usage examples, (ii) the absolute position of the target word in the examples, and (iii) the target word's relative position in the examples. Tables~\ref{tab:correlations-wordnet} and \ref{tab:correlations-oxford} show the correlation coefficients.


\begin{table}
    \centering
    \resizebox{0.5\textwidth}{!}{%
    \begin{tabular}{lrrrrrr}
        \toprule
        {} &    Length &  Relative Position &  Absolute Position &  BertScore &      Bleu &      Nist \\
        \midrule
        Length            &  1.000000 &          -0.121793 &           0.575304 &   0.067180 &  0.076133 &  0.044873 \\
        Relative Position & -0.121793 &           1.000000 &           0.626032 &   0.052725 &  0.074697 &  0.062041 \\
        Absolute Position &  0.575304 &           0.626032 &           1.000000 &   0.128785 &  0.159078 &  0.110559 \\
        BertScore         &  0.067180 &           0.052725 &           \textbf{0.128785} &   1.000000 &  0.121067 &  0.095343 \\
        Bleu              &  0.076133 &           0.074697 &           \textbf{0.159078} &   0.121067 &  1.000000 &  0.821956 \\
        Nist              &  0.044873 &           0.062041 &           \textbf{0.110559} &   0.095343 &  0.821956 &  1.000000 \\
        \bottomrule
        \end{tabular}
    }
    \caption{Correlations between properties of the usage examples and the quality (BertScore, BLEU, NIST) of the definitions generated by Flan-T5 Base for WordNet. The prompt used is `What is the definition of $w$?' (post). The maximum context size is set to 512.}
    \label{tab:correlations-wordnet}
\end{table}

\begin{table}
    \centering
    \resizebox{0.5\textwidth}{!}{%
    \begin{tabular}{lrrrrrr}
        \toprule
        {} &    Length &  Relative Position &  Absolute Position &  BertScore &      Bleu &      Nist \\
        \midrule
        Length            &  1.000000 &          -0.040948 &           0.615536 &   0.019844 &  0.039525 &  0.017253 \\
        Relative Position & -0.040948 &           1.000000 &           0.674509 &   0.046071 &  0.019940 &  0.023542 \\
        Absolute Position &  0.615536 &           0.674509 &           1.000000 &   0.029413 &  0.016901 &  0.006764 \\
        BertScore         &  0.019844 &           0.046071 &           0.029413 &   1.000000 &  0.283203 &  0.276626 \\
        Bleu              &  0.039525 &           0.019940 &           0.016901 &   0.283203 &  1.000000 &  0.687382 \\
        Nist              &  0.017253 &           0.023542 &           0.006764 &   0.276626 &  0.687382 &  1.000000 \\
        \bottomrule
        \end{tabular}
    }
    \caption{Correlations between properties of the usage examples and the quality (BertScore, BLEU, NIST) of the definitions generated by Flan-T5 Base for Oxford. The prompt used is `What is the definition of $w$?' (post). The maximum context size is set to 512.}
    \label{tab:correlations-oxford}
\end{table}

\section{Prompt Selection}
\label{sec:app-prompt-selection}
As briefly discussed in Section~\ref{sec:method}, in preliminary experiments, we use the pretrained Flan-T5 Base model \cite[250M
parameters;][]{chung2022scaling} to select a definition generation prompt among 8 alternative verbalisations. These are a combination of four different instruction strings (`Define $w$', `Define the word $w$', `Give the definition of $w$', `What is the definition of $w$?) and two ways of concatenating instructions to usage examples -- i.e., either prepending them or appending them. Tables~\ref{tab:prompt-selection-wordnet}-\ref{tab:prompt-selection-codwoe-trial} show the results of our experiments. In the tables, the strings `pre' and `post' refer to the concatenation method (prepending or appending the instruction), the numbers 128, 256, and 512 refer to the maximum length of the usage examples provided to Flan-T5 (in sub-words), and `filter' refers to the decoding strategy of always avoiding the target word (definiendum).

\begin{table}
\centering
\resizebox{0.5\textwidth}{!}{%
\begin{tabular}{lrrr}
\toprule
                                    Configuration &   BLEU &   NIST &  BERTScore \\
\midrule
what is the definition of <trg>? post 256 & 0.0985 & 0.1281 &     0.8700 \\
what is the definition of <trg>? post 512 & 0.0985 & 0.1281 &     0.8700 \\
 give the definition of <trg> post filter & 0.0719 & 0.1520 &     0.8560 \\
    give the definition of <trg> post 256 & 0.0629 & 0.1563 &     0.8522 \\
    give the definition of <trg> post 512 & 0.0629 & 0.1563 &     0.8522 \\
           define the word <trg> post 512 & 0.0462 & 0.0972 &     0.8512 \\
           define the word <trg> post 256 & 0.0462 & 0.0972 &     0.8512 \\
    give the definition of <trg>: pre 256 & 0.0446 & 0.1123 &     0.8495 \\
 what is the definition of <trg>? pre 512 & 0.0403 & 0.0705 &     0.8495 \\
    give the definition of <trg>: pre 512 & 0.0446 & 0.1123 &     0.8495 \\
 what is the definition of <trg>? pre 256 & 0.0403 & 0.0703 &     0.8494 \\
           define the word <trg>: pre 512 & 0.0313 & 0.0615 &     0.8481 \\
           define the word <trg>: pre 256 & 0.0313 & 0.0618 &     0.8480 \\
                    define <trg> post 512 & 0.0275 & 0.0583 &     0.8475 \\
                    define <trg> post 256 & 0.0275 & 0.0583 &     0.8475 \\
                    define <trg>: pre 512 & 0.0195 & 0.0411 &     0.8453 \\
                    define <trg>: pre 256 & 0.0195 & 0.0409 &     0.8453 \\
\bottomrule
\end{tabular}
}
\caption{Prompt selection results on WordNet (see description in Appendix~\ref{sec:app-prompt-selection}).}
\label{tab:prompt-selection-wordnet}
\end{table}

\begin{table}
\centering
\resizebox{0.5\textwidth}{!}{%
\begin{tabular}{lrrr}
\toprule
                                          Configuration &   BLEU &   NIST &  BERTScore \\
\midrule
       what is the definition of <trg>? post 512 & 0.1232 & 0.1488 &     0.8648 \\
       what is the definition of <trg>? post 128 & 0.1232 & 0.1488 &     0.8648 \\
       what is the definition of <trg>? post 256 & 0.1232 & 0.1488 &     0.8648 \\
what is the definition of <trg>? post oxford filter 128 & 0.1219 & 0.1398 &     0.8644 \\
           give the definition of <trg> post 128 & 0.0823 & 0.1793 &     0.8531 \\
           give the definition of <trg> post 256 & 0.0823 & 0.1793 &     0.8531 \\
           give the definition of <trg> post 512 & 0.0823 & 0.1793 &     0.8531 \\
    give the definition of <trg> post oxford filter 128 & 0.0763 & 0.1415 &     0.8526 \\
        what is the definition of <trg>? pre 256 & 0.0801 & 0.0966 &     0.8501 \\
        what is the definition of <trg>? pre 512 & 0.0801 & 0.0966 &     0.8501 \\
        what is the definition of <trg>? pre 128 & 0.0801 & 0.0966 &     0.8501 \\
           give the definition of <trg>: pre 128 & 0.0695 & 0.1313 &     0.8493 \\
           give the definition of <trg>: pre 256 & 0.0695 & 0.1313 &     0.8493 \\
           give the definition of <trg>: pre 512 & 0.0695 & 0.1313 &     0.8492 \\
                  define the word <trg> post 128 & 0.0614 & 0.1112 &     0.8442 \\
                  define the word <trg> post 512 & 0.0614 & 0.1112 &     0.8442 \\
                  define the word <trg> post 256 & 0.0614 & 0.1112 &     0.8442 \\
                  define the word <trg>: pre 256 & 0.0408 & 0.0602 &     0.8352 \\
                  define the word <trg>: pre 512 & 0.0408 & 0.0602 &     0.8352 \\
                  define the word <trg>: pre 128 & 0.0408 & 0.0602 &     0.8352 \\
                           define <trg> post 256 & 0.0279 & 0.0581 &     0.8319 \\
                           define <trg> post 128 & 0.0279 & 0.0581 &     0.8319 \\
                           define <trg> post 512 & 0.0279 & 0.0581 &     0.8319 \\
                           define <trg>: pre 512 & 0.0161 & 0.0237 &     0.8305 \\
                           define <trg>: pre 256 & 0.0160 & 0.0237 &     0.8305 \\
                           define <trg>: pre 128 & 0.0160 & 0.0237 &     0.8305 \\
\bottomrule
\end{tabular}
}
\caption{Prompt selection results on Oxford (see description in Appendix~\ref{sec:app-prompt-selection}).}
\label{tab:prompt-selection-oxford}
\end{table}

\begin{table}
\centering
\resizebox{0.5\textwidth}{!}{%
\begin{tabular}{lrrr}
\toprule
                                   Configuration &   BLEU &   NIST &  BERTScore \\
\midrule
what is the definition of <trg>? post 128 & 0.1138 & 0.2137 &     0.8702 \\
    give the definition of <trg> post 128 & 0.0826 & 0.2389 &     0.8615 \\
 what is the definition of <trg>? post 64 & 0.1033 & 0.1990 &     0.8595 \\
     give the definition of <trg> post 64 & 0.0785 & 0.2194 &     0.8520 \\
\bottomrule
\end{tabular}
}
\caption{Prompt selection results on CoDWoE Complete (see description in Appendix~\ref{sec:app-prompt-selection}).}
\label{tab:prompt-selection-codwoe-complete}
\end{table}

\begin{table}
\centering
\resizebox{0.5\textwidth}{!}{%
\begin{tabular}{lrrr}
\toprule
                                        Configuration &   BLEU &   NIST &  BERTScore \\
\midrule
    give the definition of <trg>: pre 64 & 0.0680 & 0.1513 &     0.8461 \\
what is the definition of <trg>? post 64 & 0.1068 & 0.1464 &     0.8458 \\
    give the definition of <trg> post 64 & 0.0654 & 0.1602 &     0.8374 \\
\bottomrule
\end{tabular}
}
\caption{Prompt selection results on CoDWoE Trial (see description in Appendix~\ref{sec:app-prompt-selection}).}
\label{tab:prompt-selection-codwoe-trial}
\end{table}

\begin{table*}
\centering
\resizebox{0.9\textwidth}{!}{%
\begin{tabular}{llcccccc}
 & &
  \multicolumn{3}{c}{\textit{\textbf{WordNet}}} &
  \multicolumn{3}{c}{\textit{\textbf{Oxford}}} 
  \\ \toprule
  Model &
  Test &
  BLEU &
  ROUGE-L &
  \multicolumn{1}{c|}{BERT-F1} &
  BLEU &
  ROUGE-L &
  \multicolumn{1}{c}{BERT-F1} 
  \\ \midrule
\citet{huang-etal-2021-definition} & \textit{Unknown} &  32.72 & - & \multicolumn{1}{c|}{-} & \textbf{26.52} & - & \multicolumn{1}{c}{-} \\ 
T5 base & Zero-shot (task shift) &
2.01 &  
8.24 & 
  \multicolumn{1}{c|}{82.98} &
1.72 &
7.48 &
  \multicolumn{1}{c}{78.79}
   \\
T5 base & Soft domain shift &
9.21 &
25.71 & 
  \multicolumn{1}{c|}{86.44} &
 7.28&
 24.13&
  \multicolumn{1}{c}{86.03} 
   \\

Flan-T5 base & Zero-shot (task shift) &
   4.08 &  
   15.32 & 
  \multicolumn{1}{c|}{87.00} &
   3.71 &
   17.25 &
  \multicolumn{1}{c}{86.44}
   \\
   Flan-T5 base & In-distribution &
   8.80 &
   23.19 & 
  \multicolumn{1}{c|}{87.49} &
   6.15 &
   20.84 &
  \multicolumn{1}{c}{86.48} 
   \\
     Flan-T5 base & Hard domain shift &
   6.89 &
   20.53 &
  \multicolumn{1}{c|}{87.16} &
   4.32 &
   17.00 &
  \multicolumn{1}{c}{85.88} 
   \\
    Flan-T5 base & Soft domain shift &
   10.38 &
   27.17 &
  \multicolumn{1}{c|}{88.22} &
   7.18 &
   23.04 &
  \multicolumn{1}{c}{86.90} 
   \\
  Flan-T5 large & Soft domain shift &
   14.37 &
   33.74 &
  \multicolumn{1}{c|}{88.21} &
   10.90 &
   30.05 &
  \multicolumn{1}{c}{87.44} 
   \\
T5 XL & Zero-shot (task shift) &
2.05 &  
8.28 & 
  \multicolumn{1}{c|}{81.90} &
2.28 &
9.73 &
  \multicolumn{1}{c}{80.37}
   \\
T5 XL & Soft domain shift &
\textbf{34.14} &
\textbf{53.55} & 
  \multicolumn{1}{c|}{91.40} &
18.82 &
38.26 &
  \multicolumn{1}{c}{88.81} 
   \\
   Flan-T5 XL & Zero-shot (task shift) &
   2.70 &  
   12.72 & 
  \multicolumn{1}{c|}{86.72} &
   2.88 &
   16.20 &
  \multicolumn{1}{c}{86.52}
   \\
   Flan-T5 XL & In-distribution &
   11.49 &
   28.96 & 
  \multicolumn{1}{c|}{88.90} &
   16.61 &
   36.27 &
  \multicolumn{1}{c}{89.40} 
   \\
  Flan-T5 XL & Hard domain shift &
   29.55 &
   48.17 &
  \multicolumn{1}{c|}{91.39} &
   8.37 &
   25.06 &
  \multicolumn{1}{c}{87.56} 
   \\
   Flan-T5 XL & Soft domain shift &
   \textbf{32.81} &
   \textbf{52.21} &
  \multicolumn{1}{c|}{\textbf{92.16}} &
  18.69 &
  \textbf{38.72} &
  \multicolumn{1}{c}{\textbf{89.75}} 
   \\ \bottomrule
\end{tabular}%
}
\caption{Results of the definition generation experiments.}
\label{tab:def-gen-results-base}
\end{table*}

\section{Additional Results}
\subsection{Zero-Shot Evaluation of Flan-T5 (Task Shift)}
\label{sec:appendix-zero-shot}
Here we directly evaluate Flan-T5 XL on the WordNet and Oxford test sets, without any fine-tuning nor in-context learning.\footnote{
We only condition generation on the usage examples and the task prompt. We do \textit{not} provide full instances (i.e., usage examples, task prompts, and definitions) in the context, as one would do in a few-shot setup.} 
Table \ref{tab:def-gen-results} in the main paper shows low BLEU and ROUGE-L scores but rather high BERT-F1. 
Overall, the model does not exhibit consistent task understanding (e.g. it generates \sense{skepticism} as a definition for \word{healthy} as exemplified in the phrase \textit{`healthy skepticism'}). A qualitative inspection, however, reveals that the generated definitions can still be often informative (e.g., \sense{a workweek that is longer than the regular workweek} is informative with respect to the meaning of \word{overtime} although the ground truth definition is \sense{beyond the regular time}).
The two surface-overlap metrics cannot capture this, but the relatively high BERT-F1 confirms that the semantic content of generations is largely appropriate.
There are indeed also many good zero-shot definitions. For example \sense{intense} for \word{fervent} as in \textit{`the fervent heat'}, or \sense{a conversation} for \word{discussion} in \textit{`we had a good discussion'}.

\subsection{Other Models and Model Variants}
\label{sec:appendix-other-models}
We evaluate T5 (base and XL) and Flan-T5 (base, large, and XL) under the same generalisation conditions presented for Flan T5 XL in the main paper (Section~\ref{sec:method-def-eval}) and above in Appendix~\ref{sec:appendix-zero-shot}. Results for FlanT5-XL are reported in the main paper (Table~\ref{tab:def-gen-results}); here, in Table~\ref{tab:def-gen-results-base}, we report results for all models and model variants.

\subsection{Evaluation Cards}
\label{sec:appendix-eval-cards}
In Table~\ref{tab:eval_card}, we provide an evaluation card
to clarify the nature of the generalisation tests performed on definition generators.\footnote{
    \url{https://genbench.org/eval_cards}
} In-distribution tests are not included as they do not include any shift between the training and test data distributions \cite{hupkes2022state}. 
We also register our work in the GenBench evolving survey of generalisation in NLP.\footnote{
    \url{https://genbench.org/references}
}\looseness-1
%
\newcommand{\tabularwidth}{\columnwidth}

\newcommand{\expone}{$\square$}
\newcommand{\exptwo}{$\bigtriangleup$}
\newcommand{\expthree}{$\bigcirc$}
        
\renewcommand{\arraystretch}{1.1}         
\setlength{\tabcolsep}{0mm}         

\begin{table}
\centering
\begin{tabular}{|p{\tabularwidth}<{\centering}|}         
\hline
               
\rowcolor{gray!60}               
\textbf{Motivation} \\               
\footnotesize
\begin{tabular}{p{0.25\tabularwidth}<{\centering} p{0.25\tabularwidth}<{\centering} p{0.25\tabularwidth}<{\centering} p{0.25\tabularwidth}<{\centering}}                        
\textit{Practical} & \textit{Cognitive} & \textit{Intrinsic} & \textit{Fairness}\\
\expone\hspace{0.8mm}\exptwo\hspace{0.8mm}\expthree\hspace{0.8mm}		
& 		
& 		
& 		

\vspace{2mm} \\
\end{tabular}\\
               
\rowcolor{gray!60}               
\textbf{Generalisation type} \\               
\footnotesize
\begin{tabular}{m{0.17\tabularwidth}<{\centering} m{0.20\tabularwidth}<{\centering} m{0.14\tabularwidth}<{\centering} m{0.17\tabularwidth}<{\centering} m{0.18\tabularwidth}<{\centering} m{0.14\tabularwidth}<{\centering}}                   
\textit{Compo- sitional} & \textit{Structural} & \textit{Cross Task} & \textit{Cross Language} & \textit{Cross Domain} & \textit{Robust- ness}\\
& 		
& \expone\hspace{0.8mm}		
& 		
& \exptwo\hspace{0.8mm}\expthree\hspace{0.8mm}		
& 		

\vspace{2mm} \\
\end{tabular}\\
             
\rowcolor{gray!60}             
\textbf{Shift type} \\             
\footnotesize
\begin{tabular}{p{0.25\tabularwidth}<{\centering} p{0.25\tabularwidth}<{\centering} p{0.25\tabularwidth}<{\centering} p{0.25\tabularwidth}<{\centering}}                        
\textit{Covariate} & \textit{Label} & \textit{Full} & \textit{Assumed}\\  
\exptwo\hspace{0.8mm}\expthree\hspace{0.8mm}		
& 		
& \expone\hspace{0.8mm}		
& 		

\vspace{2mm} \\
\end{tabular}\\
             
\rowcolor{gray!60}             
\textbf{Shift source} \\             
\footnotesize
\begin{tabular}{p{0.25\tabularwidth}<{\centering} p{0.25\tabularwidth}<{\centering} p{0.25\tabularwidth}<{\centering} p{0.25\tabularwidth}<{\centering}}                          
\textit{Naturally occurring} & \textit{Partitioned natural} & \textit{Generated shift} & \textit{Fully generated}\\
\expone\hspace{0.8mm}\exptwo\hspace{0.8mm}\expthree\hspace{0.8mm}		
& 		
& 		
& 		

\vspace{2mm} \\
\end{tabular}\\
             
\rowcolor{gray!60}             
\textbf{Shift locus}\\             
\footnotesize
\begin{tabular}{p{0.25\tabularwidth}<{\centering} p{0.25\tabularwidth}<{\centering} p{0.25\tabularwidth}<{\centering} p{0.25\tabularwidth}<{\centering}}                         
\textit{Train--test} & \textit{Finetune train--test} & \textit{Pretrain--train} & \textit{Pretrain--test}\\
& \exptwo\hspace{0.8mm}\expthree\hspace{0.8mm}		
& 		
& \hspace{3em}\expone\hspace{0.8mm}		

\vspace{2mm} \\
\end{tabular}\\

\hline
\end{tabular}
\caption{Evaluation card for the generalisation tests performed on definition generators. The setups are: zero-shot (\expone), hard domain shift (\exptwo), and soft domain shift (\expthree). In-distribution tests are not included as they do not include any shift between the training and test data distributions.}
\label{tab:eval_card}
\end{table}

\section{Additional Examples of Generated Definitions and Sense Labels}
\label{sec:examples}
Some definitions generated by Flan-T5 XL manage to capture very subtle aspects of the contextual lexical meaning. In the following list, we give the usage and then the  contextual definition of \word{word}:
\begin{enumerate}
\itemsep=0em
    \item \textit{`There are people out there who have never heard of the Father, Son and Holy Spirit, let alone the \textbf{Word} of God.'}: \sense{The Bible}
    \item \textit{`Good News Bible Before the world was created, the \textbf{Word} already existed; he was with God, and he was the same as God.'}: \sense{( christianity ) the second person of the trinity ; je}
    \item \textit{`It was in that basement that I learned the skills necessary to succeed in the difficult thespian world-specifically, get up on stage, say my \textbf{words}, get off the stage-skills...'}: \sense{The dialogue of a play.}
\end{enumerate}

\begin{figure*}
    \centering
    \includegraphics[width=0.45\linewidth]{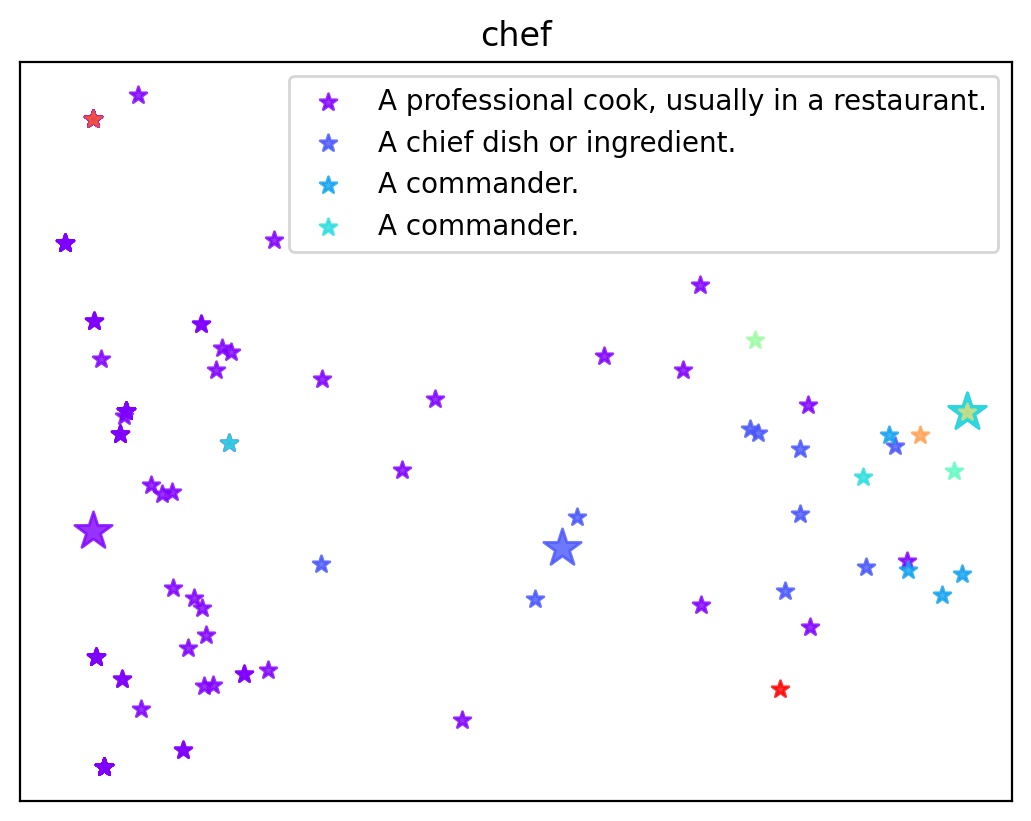}
    \includegraphics[width=0.45\linewidth]{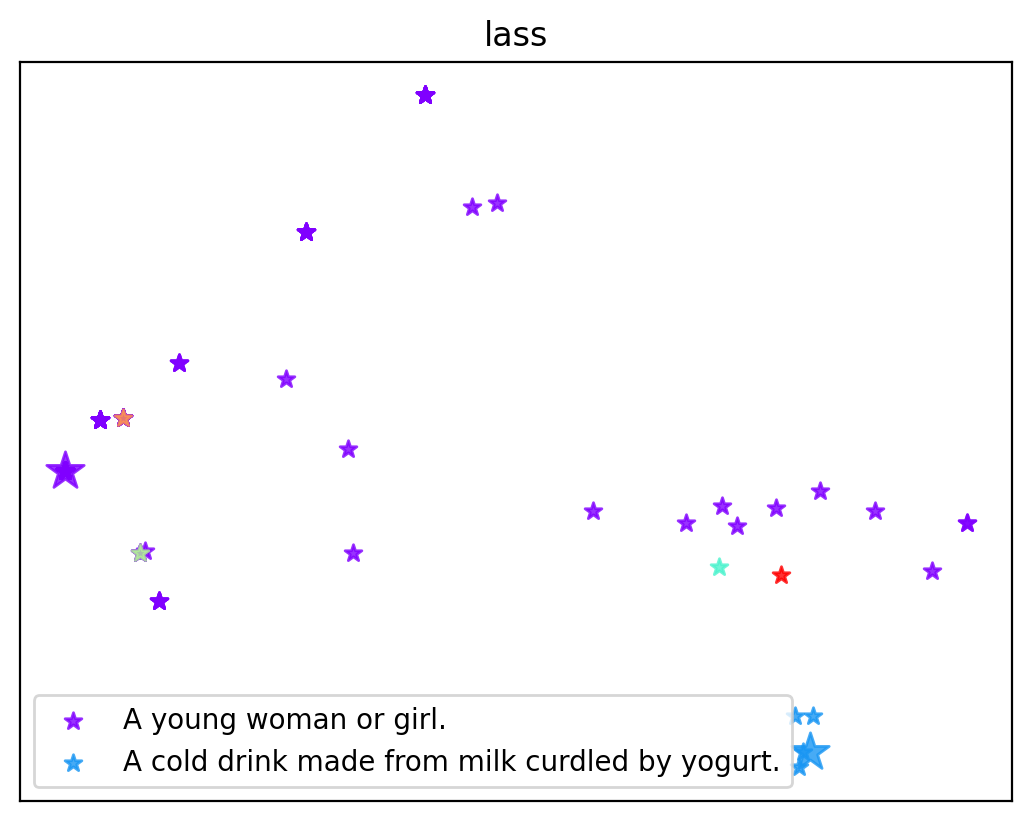}
    \caption{PCA projections of definition embeddings for two target words from English DWUG.}
    \label{fig:sense_labels}
\end{figure*}

Interesting insights can be drawn from how the embeddings of the generated definitions are located in the vector space. 
Figure~\ref{fig:sense_labels} shows PCA projections of definition embeddings for usages of the words \word{chef} and \word{lass} from the English DWUG. Colours represent sense clusters provided in the DWUG, and the legend shows most prototypical definitions for each sense generated by our best system (singleton clusters are ignored). The large star for each sense  corresponds to its sense label (as opposed to smaller stars corresponding to other definitions not chosen as the label).

For the word \word{chef}, there are two sense clusters, for which an identical definition is chosen (\sense{A commander}). This most probably means that these clusters should in fact be merged together, or that they are in the process of splitting (see also Section~\ref{sec:explain}). These two senses are (not surprisingly) much closer to each other than to the definitions from the \sense{professional cook} sense.
For the word \word{lass}, it is interesting how  separate is a small bluish group of definitions in the bottom right corner of the plot, where the target form is actually \word{lassi}. The fine-tuned Flan-T5-XL model defined this group as \sense{A cold drink made from milk curdled by yogurt}, which is indeed what \word{lassi} is (ignoring minor details).

\section{Human Evaluation Guidelines}
\label{sec:guidelines}

\begin{figure}
    \centering
    \includegraphics[width=\linewidth]{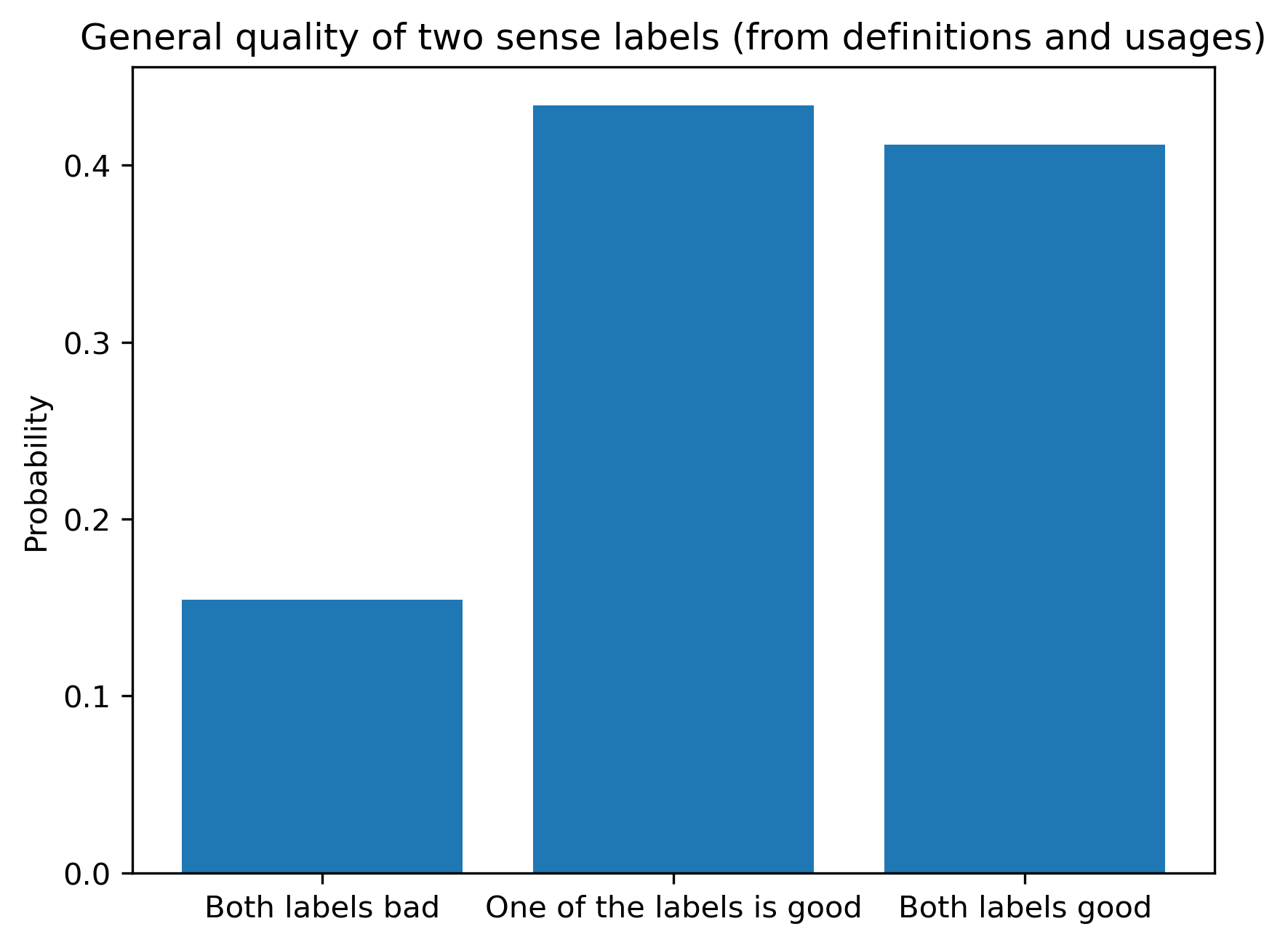}
    \caption{General quality of generated sense labels}
    \label{fig:annot_quality}
\end{figure}
\begin{figure}
    \centering
    \includegraphics[width=\linewidth]{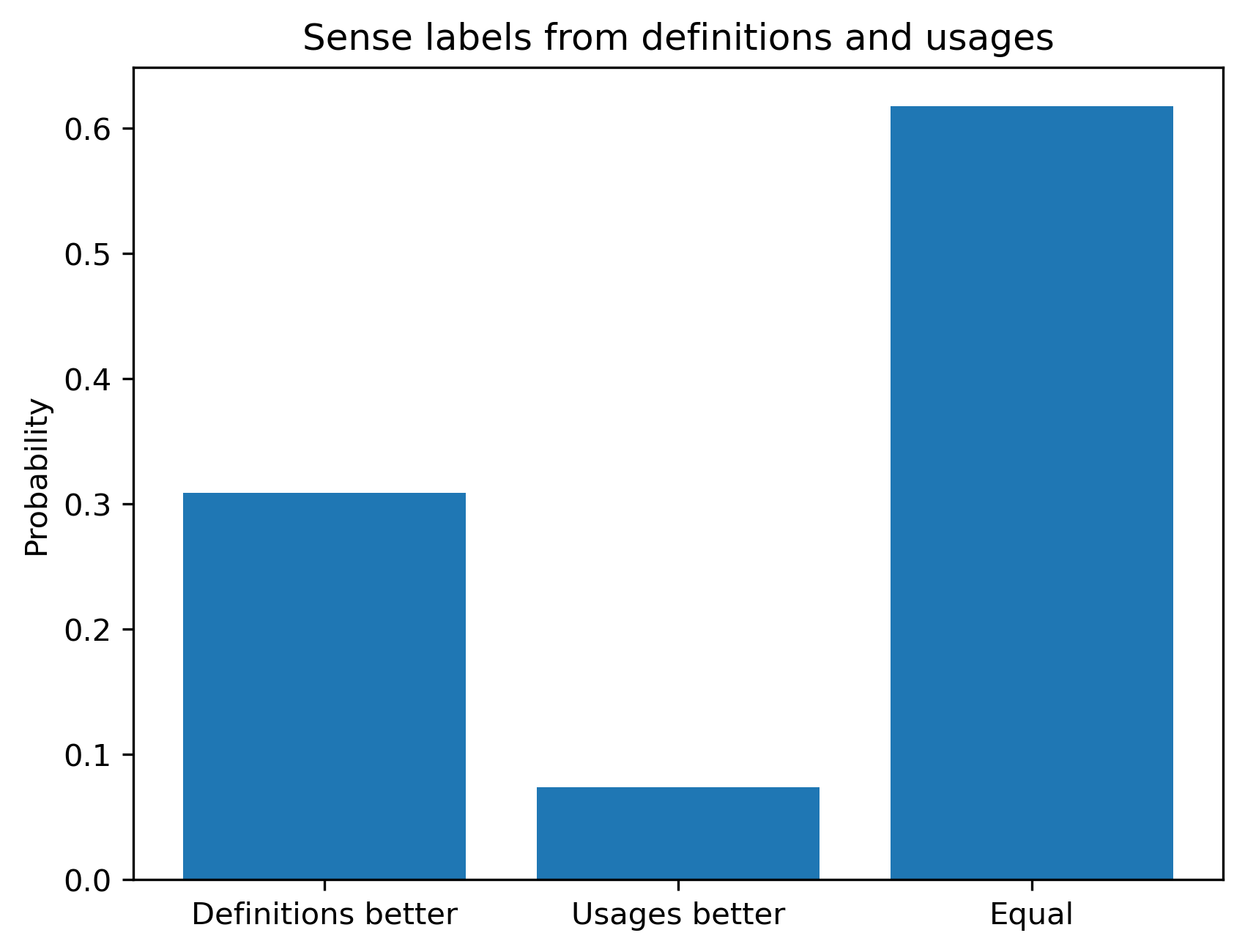}
    \caption{Human comparison of sense labels produced from definitions and from usages}
    \label{fig:definitions_better}
\end{figure}

\textit{Figures \ref{fig:annot_quality} and \ref{fig:definitions_better} show the results of the human evaluation.}

`You are given a spreadsheet with four columns: \textbf{Targets}, \textbf{Examples}, \textbf{System1} and \textbf{System2}. In every row, we have one target English word in the \textbf{Targets} column and five (or less) example usages of this word in the Examples column. Usages are simply sentences with at least one occurrence of the target word: one usage per line.

Every row is supposed to contain usages where the target word is used in the same sense: this means that for ambiguous words, there will be multiple rows, each corresponding to a particular sense. This division into senses is not always 100\% correct, but for the purposes of this annotation effort, we take it for granted. Note that the five example usages in each row are sampled randomly from a larger set of usages belonging to this sense.

System1 and System2 are computational models which produce human-readable labels or definitions for each sense of a target word. They employ different approaches, and your task is to compare and evaluate the labels generated by these two systems. Note that in each row, the names `System1' and `System2' are randomly assigned to the actual generation systems.

The  generated sense labels are supposed to be useful for historical linguists and lexicographers. Thus, they must be:
\begin{enumerate}
    \item \textbf{Truthful}: i.e., should reflect exactly the sense in which the target word is occurring in the example usages. Ideally, the label should be general enough to encompass all the usages from the current row, but also specific enough so as not to mix with other senses (for poly-semantic target words).
    \item \textbf{Fluent}: i.e., feeling like natural English sentence or sentences, without grammar errors, utterances broken mid-word, etc
\end{enumerate}

You have to fill in the \textbf{Judgements} column with one of six integer values:
\begin{itemize}
    \item \textbf{0}: both systems are equally bad for this sense
    \item \textbf{1}: System 1 is better, but System 2 is also OK
    \item \textbf{11}: System 1 is better, and System 2 is bad
    \item \textbf{2}: System 2 is better, but System 1 is also OK
    \item \textbf{22}: System 2 is better, and System 1 is bad
    \item \textbf{3}: both systems are equally good for this sense
\end{itemize}

Some rows are already pre-populated with the \textbf{3} judgement, because the sense labels generated by both systems are identical. We hypothesise that this most probably means that both labels are equally good. Please still have a look at these identical labels and change \textbf{3} to \textbf{0} in case you feel that in fact they are equally bad.'

\section{Sense Dynamics Maps}
\label{sec:appendix-maps}
It is easy to find different sense clusters which are assigned \textit{identical} definition labels. 
Usage examples from sense clusters $c_2$ and $c_3$ for the word \word{chef}, to which our system assigned the same label: \sense{A commander}:
\begin{itemize}
    \item $c_2$: `\textit{He boasted of having been a \textbf{chef} de brigade in the republican armies of France}', `\textit{Morrel has received a regiment, and Joliette is \textbf{Chef} d'Escadron of Spahis}', `\textit{as major-general and \textbf{chef} d'escadron, during the pleasure of our glorious monarch Louis le Grand}'
    \item $c_3$: `\textit{That brave general added to his rank of \textbf{chef} de brigade that of adjutant general}', `\textit{I frequently saw Mehevi and several other \textbf{chefs} and warriors of note take part}'
\end{itemize}
Thus, a user can safely accept the suggestion of our system to consider these two clusters as one sense.

Note that \sense{A commander} practically disappeared as a word sense in the 20th century, replaced by \sense{a professional cook, usually in a restaurant}.

\section{Clustering Embedding Spaces}
\label{sec:appendix-embedding-spaces}

We constructed three types of embedding spaces; (i) contextualised token embeddings, (ii) sentence embeddings, and (ii) definition embeddings. We did so for two language models: RoBERTa-large and DistilRoBERTa. Since we cluster the embedding spaces for each target word individually, we obtain different optimal number of clusters for each target word. Table \ref{tab:Space_averages} displays the average results over all target words. 

We observe that the optimal number of clusters $K$ is substantially higher for the definition embedding spaces for both RoBERTa-large and DistilRoBERTa. However, this is an artefact of the data: since some distinct usages yield identical definitions for a target word, the definition space oftentimes consist of less distinct data points, which greatly impacts the average silhouette scores. Future work should point out what clustering methods are most applicable to definition embedding spaces. Still, this decrease in data points confirms how the definition embedding space could represent usages at a higher level of abstraction, collapsing distinct usages into identical representations. 

\begin{figure}[ht]
    \centering
    \includegraphics[trim={7cm 2cm 7cm 1cm},clip, width=\linewidth]{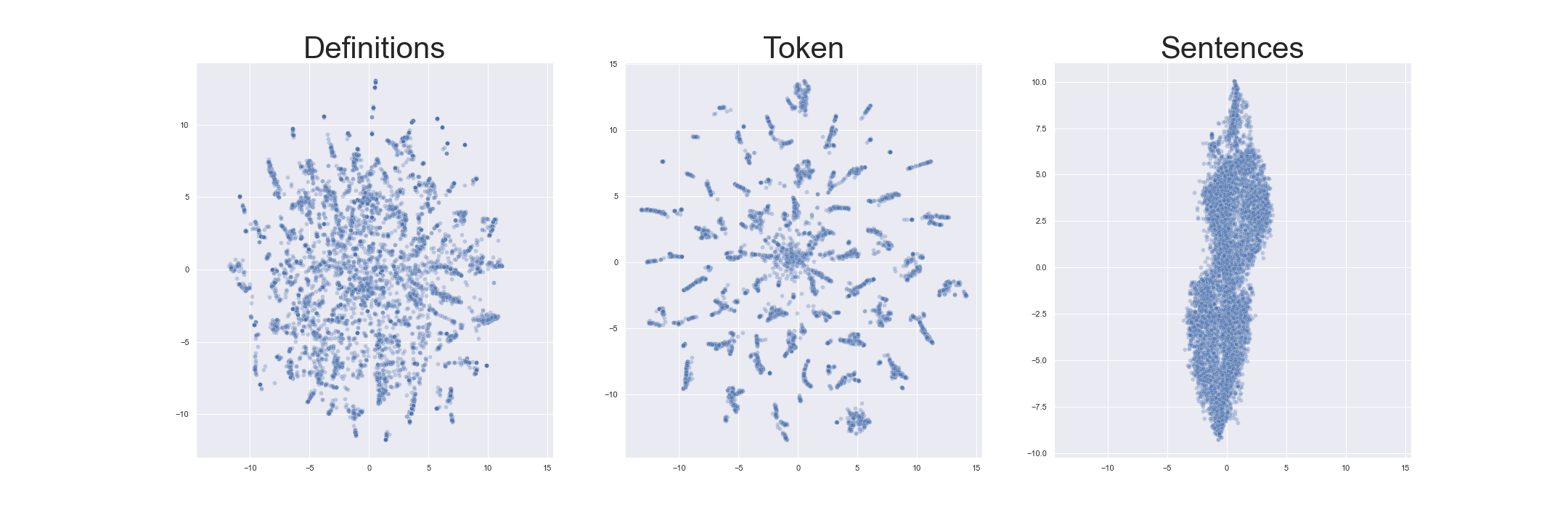}
    \caption{T-SNE projection of each embedding space, RoBERTa-Large model.}
    \label{fig:T-SNE-roberta}
\end{figure}

Figure \ref{fig:T-SNE-roberta} displays the T-SNE projections of each of the three embedding spaces of RoBERTA-large. As for Distil-RoBERTa, the definition embedding space appears to have spacial properties that are more similar to contextualised \textit{token} embedding spaces than to \textit{sentence} embedding spaces: the definition embeddings are more separated than the sentence embeddings, and are cluttered in a similar manner as the token embeddings.


Table~\ref{tab:DWUG_averages} shows the average inter- and intra-cluster dispersion values of the clusters as labelled by the English DWUGs \cite{schlechtweg-etal-2021-dwug}. These are calculated for the token, sentence and definition embeddings of both RoBERTa-large and Distil-RoBERTa.

\begin{table}
\resizebox{\columnwidth}{!}{%
\begin{tabular}{l@{\hspace{0.2cm}}l@{\hspace{0.2cm}}r@{\hspace{0.2cm}}r@{\hspace{0.2cm}}r}
\toprule
\textbf{Model} & \textbf{Representation} & \textbf{Sep.} $\uparrow$ & \textbf{Coh.} $\downarrow$ & \textbf{Ratio} $\uparrow$\\
\midrule
          & Sentence                   &    0.017 &  0.013 &  1.248 \\
RoBERTa-large                       & Token                      &    0.042 &  0.034 &  1.272 \\
                       & Definitions                &    0.008 &  0.006 &  1.349 \\ \midrule
          & Sentence                   &    0.665 &  0.592 &  1.126 \\
DistilRoBERTa                       & Token                      &    0.591 &  0.477 &  1.258 \\
                       & Definitions                &    0.705 &  0.509 &  1.397 \\
\bottomrule
\end{tabular}
}
\caption{Separation score, cohesion score, and separation-cohesion ratio for each embedding space; average over all target words from the English DWUGs.}
\label{tab:DWUG_averages}
\end{table}

\end{document}